%% file: root.tex
\documentclass[letterpaper, 10 pt, conference]{ieeeconf}  
\usepackage[utf8]{inputenc}
\usepackage[T1]{fontenc}
\usepackage{mathtools}
\usepackage{booktabs}
\usepackage{xcolor}
\usepackage{pifont}
\usepackage{times}
\usepackage{amsmath, amssymb, bm}
\usepackage{graphicx}
\usepackage{cite}
\usepackage{multirow}
\usepackage{needspace}

\makeatletter
\let\JournalCaption\@makecaption
\makeatother
\usepackage[compatibility=false]{caption}
\usepackage{subcaption}
\makeatletter
\let\@makecaption\JournalCaption
\makeatother

\allowdisplaybreaks
\usepackage{algorithm}
\usepackage{algpseudocode}

\IEEEoverridecommandlockouts                              

\title{\LARGE \bf
$\text{SG}^2$SLAM: Spatial Geometry-Informed 3D Gaussian Splatting SLAM for Monocular Outdoor Navigation
}

\author{Ziang Guo$^{\dagger}$, Didar Seyidov$^{\dagger}$, Yixiao Zhou$^{\S}$, Xuefeng Zhang$^{\ddag}$, \\ Ali Alridha Abdulkarim$^{\dagger}$, Chen Min$^{\circ*}$, Zufeng Zhang$^{\ddag*}$, and Dzmitry Tsetserukou$^{\dagger}$
\thanks{$^{*}$ are the corresponding authors.}
\thanks{$^{\dagger}$Ziang Guo, Didar Seyidov, Ali Alridha Abdulkarim, and Dzmitry Tsetserukou are with the Intelligent Space Robotics Laboratory, Skolkovo Institute of Science and Technology, Moscow, Russia., {\tt\small \{ziang.guo, Didar.Seyidov, ali.abdulkarim, d.tsetserukou\}@skoltech.ru}}%
\thanks{$^{\circ}$Chen Min is with the Research Center for Intelligent Computing Systems, SKLP, Institute of Computing Technology, Chinese Academy of Sciences, Beijing, China., {\tt\small mincheng@ict.ac.cn}}%
\thanks{$^{\S}$Yixiao Zhou is with the Department of Electrical and Electronic Engineering, The University of Hong Kong, China., {\tt\small u3649491@connect.hku.hk}}%
\thanks{$^{\ddag}$Xuefeng Zhang is with SuZhou Automotive Research Institute, Tsinghua University, China. Zufeng Zhang is with SuZhou Automotive Research Institute, Tsinghua University and the School of Vehicle and Mobility, Tsinghua University, China., {\tt\small osu\_zxf@126.com, zhangzufeng@tsari.tsinghua.edu.cn}%
}}

\begin{document}

\maketitle
\thispagestyle{empty}
\pagestyle{empty}

\begin{abstract}

Monocular outdoor navigation requires robust camera motion estimation and a dense, navigable map, where 3D Gaussian Splatting SLAM serves this task through an explicit differentiable parameterization and closed-form projection. However, typically, the geometry is recovered by associating 2D pixels with 3D points, so a modular SLAM system absorbs tracking error into the map, leading to cascading degradation. To address these gaps, we present $\text{SG}^2$SLAM, which couples streaming spatial geometry with a ray-conditioned 3DGS kernel. Streaming spatial geometry replaces the per-frame cross-dimensional correspondence with a chunk-level metric prior that supervises tracking and mapping from one source. A ray-conditioned kernel extends each primitive with a viewing-direction subspace that senses the monocular camera viewpoints. The joint optimization over spatial geometry and poses ensures a globally consistent map and trajectory outcome. Experiments on the KITTI and Waymo Open datasets show that our $\text{SG}^2$SLAM surpasses the per-frame association baselines in the monocular outdoor navigation task. Code will be made publicly available upon acceptance.

\end{abstract}


\section{Introduction}

\input{Intro}

\section{Related Work}

\input{Related}

\section{Methodology}

\input{Method}

\section{Experimental Results}

\input{Exp}

\section{Conclusion}

This paper presents $\text{SG}^2$SLAM, a 3DGS SLAM system built on streaming spatial geometry, which dispenses with the per-frame 2D–3D association and pairwise image matching. Our system introduces a novel paradigm that boosts monocular 3DGS SLAM by enabling joint 3D–3D association for both mapping and tracking in a streaming manner. Experiments on the KITTI, our own real-world driving, and Waymo Open Datasets show that our $\text{SG}^2$SLAM exceeds existing paradigms in monocular outdoor navigation. Future work will focus on the robustness of perception quality under unstructured conditions, e.g., rainy nights and wild forests, to further improve the stability of our system.


\bibliographystyle{IEEEtran}
\bibliography{refs}
\end{document}

%% file: Intro.tex
Recovering camera motion and creating a dense, navigable map of the environment from a monocular RGB stream remain core capabilities for autonomous navigation. Coupling 3D Gaussian Splatting (3DGS) with visual Simultaneous Localization and Mapping (SLAM) has rapidly become a mainstream research direction, as its explicit, differentiable parameterization is naturally compatible with online tracking and mapping \cite{patle2025ad-gs}. However, in monocular outdoor navigation, which is the deployment scenario of practical interest, the 3DGS SLAM suffers from an undetermined global scale and further exposes two concrete gaps that we identify below.

Existing SLAM systems recover geometry through the 2D–3D matching paradigm, in which 2D pixel observations are associated with 3D points \cite{deng2025gigaslam, feldmann2026diskchungs, murai2025mast3rslam, yu2026kilogs, cheng2025outdoor}. The 2D–3D association is, however, intrinsically cross-dimensional. The 3D to 2D projection irreversibly discards depth, and the inverse recovery is a non-convex problem that is sensitive to outliers, reprojection noise, and mismatched feature scales. Under the conditions of outdoor monocular navigation, these sensitivities compound. Within a modular SLAM architecture, tracking errors from 2D–3D correspondences are absorbed into the 3D map, which then provides a degraded reference for the next round of matching, producing a slow but unbounded drift that is particularly hard to localize and correct.

The 3DGS map becomes geometrically ambiguous while absorbing tracking errors. 3DGS is mainly optimized against a photometric loss for novel view synthesis, which admits multiple geometrically distinct Gaussian configurations that produce indistinguishable renderings. In outdoor sequences, this manifests as floating Gaussians, broken thin structures, and inconsistent surface normals \cite{huang20242dgs}. These artifacts are tolerable for view synthesis but are directly harmful for navigation due to the pollution of depth information, where downstream tasks such as collision checking and traversability estimation consume the map.

To mitigate these gaps in outdoor monocular conditions, we propose $\text{SG}^2$SLAM, a 3DGS SLAM built on streaming spatial geometry and a monocular-depth-aware ray-conditioned 3DGS kernel, where the frontend streams poses, point maps, and per-pixel confidence from windows of frames to form chunking, with a metric depth network supervising the metric scale. Then, the backend refines the 3DGS map, conditioning each primitive on the viewing ray with spatial geometric supervision and keeping the mapping consistent and aware under monocular camera conditions. Chunks are then fused by a joint global graph optimization over spatial geometry and poses. Our contribution is as follows:
\begin{itemize}
    \item We present a 3DGS SLAM built on streaming spatial geometry. Compared to the baselines based on pairwise 2D pixel matching and cross-dimensional correspondences, our $\text{SG}^2$SLAM improves both frontend tracking and backend mapping.

    \item We render 3DGS through a ray-conditioned kernel in closed form, where the mean and covariance span a spatial block and a viewing-direction subspace, thus removing the blending at grazing angles that may produce floating primitives.

    \item We fuse chunks through joint global graph optimization over the spatial geometry and poses, which keeps the map and trajectory globally consistent and settles the fusion of all chunks in a single solve.
\end{itemize}

%% file: Related.tex
\subsection{Classical and Learning-Based Monocular SLAM}

MonoSLAM~\cite{davison2007monoslam} showed that a single camera is enough to track and map in real time with an extended Kalman filter, and PTAM~\cite{klein2007ptam} made keyframe-based bundle adjustment practical by running tracking and mapping in two parallel threads. The ORB-SLAM family~\cite{murartal2015orbslam, murartal2017orbslam2, campos2021orbslam3} was built on this design using ORB features, a covisibility graph, and bag-of-words place recognition~\cite{galvez2012bags}. Methods such as LSD-SLAM~\cite{engel2014lsd} skip keypoints and minimize photometric error. DROID-SLAM~\cite{teed2021droid} iterates dense optical flow updates through a differentiable bundle adjustment layer and generalizes well across datasets. DPVO~\cite{teed2023dpvo} replaces dense flow with sparse patches to cut memory and runtime. All of these systems recover depth indirectly by associating 2D observations with 3D points, and the quality of representation determines both the accuracy and the cost of the association.

\subsection{SLAM with 3D Foundation Models}

Another direction uses 3D foundation models that regress 3D structures directly from images. DUSt3R~\cite{wang2024dust3r} predicts point maps for an image pair in a shared frame, without known intrinsics, and MASt3R~\cite{leroy2024mast3r} adds a matching head that provides dense and accurate correspondences. Spann3R~\cite{wang2025spann3r} processes images pairwise and maintains a spatial memory in a recurrent state, allowing the whole video to be reconstructed online. VGGT~\cite{wang2025vggt} goes further and predicts cameras, depth maps, and point maps for multiple views in a single feed-forward pass. Several SLAM systems are built on these models. MASt3R-SLAM uses two-view MASt3R predictions for tracking and local fusion, and it runs in real time on uncalibrated video \cite{murai2025mast3rslam, leroy2024mast3r}. SLAM3R~\cite{liu2025slam3r} reconstructs dense scenes from monocular RGB with a sliding window of frames that are registered into a global map. VGGT-SLAM builds submaps with VGGT and aligns them with a 15 degree of freedom homography on the SL(4) manifold, since an uncalibrated reconstruction is only defined up to a projective transformation \cite{maggio2025vggtslam, wang2025vggt}. VGGT-Long~\cite{deng2025vggtlong} targets kilometer-scale driving sequences by splitting the video into overlapping chunks, aligning neighboring chunks, and correcting the result with loop closure. This direction replaces correspondence with regression, predicting the 3D structures directly by the network rather than by the camera model. But such systems recover geometric similarity only after submap registration, sliding windows, or explicit alignment in a pairwise manner.

\subsection{3D Gaussian Splatting SLAM}

3D Gaussian Splatting~\cite{kerbl20233dgs} represents the scene with explicit anisotropic Gaussians that are rasterized at high frame rates, and it quickly became the preferred map for dense SLAM. MonoGS~\cite{matsuki2024gaussian} optimizes camera poses directly against the Gaussian map and works with monocular or RGB-D input. SplaTAM~\cite{keetha2024splatam} and GS-SLAM~\cite{yan2024gsslam} rely on depth sensors, while Photo-SLAM~\cite{huang2024photoslam} and CaRtGS~\cite{feng2024cartgs} leave tracking to ORB-SLAM3 and spend the GPU budget on mapping. Splat-SLAM~\cite{sandstrom2024splat} and HI-SLAM2~\cite{zhang2025hislam2} pair a learned dense tracker with monocular depth priors, and they deform the 3D Gaussians when the trajectory is corrected by global optimization. LoopSplat~\cite{zhu2025loopsplat} closes loops by registering submaps made of 3D Gaussians directly. S3PO-GS~\cite{cheng2025outdoor} anchors MASt3R point maps to the Gaussian map, which keeps the scale consistent across frames and limits scale drift. GigaSLAM~\cite{deng2025gigaslam} stores 3D Gaussians in a hierarchical sparse voxel map, tracks with a metric depth network together with epipolar geometry, and corrects drift with bag-of-words loop closure. DiskChunGS~\cite{feldmann2026diskchungs} boosts performance by optimizing the I/O: it splits the scene into spatial chunks, keeps only the active ones on the GPU, and moves the rest to disk. KiloGS-SLAM~\cite{yu2026kilogs} combines a multi-tier tracker with chunk-based densification and pruning of 3D Gaussians. Across these systems, the 3DGS map is dominated by image evidence rather than spatial geometry, so the cross-dimensional cost of 2D–3D matching 
persists beneath the recent progress.

%% file: Method.tex

\begin{figure*}
    \centering
    \includegraphics[width=0.98\linewidth]{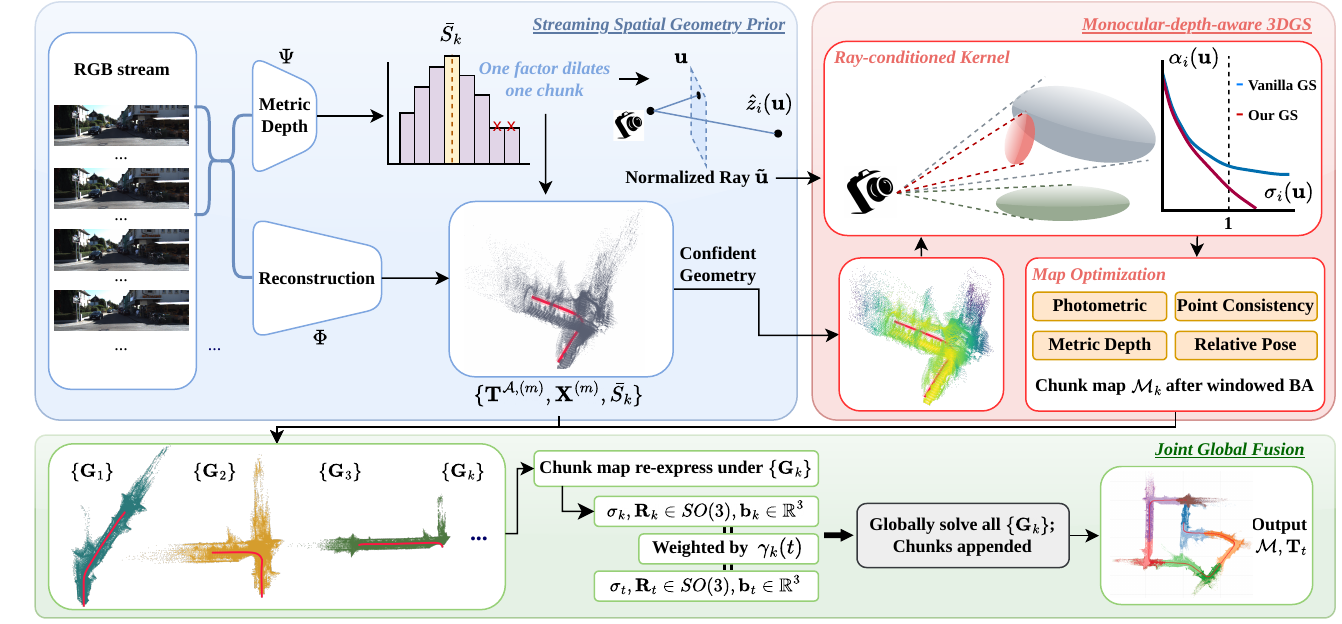}
    \vspace{-3mm}
    \caption{$\text{SG}^2$SLAM system. Streaming spatial geometry prior breaks the limitations on a per-frame 2D-to-3D association, optimizing the drift accumulation from cross-dimensional association. 3DGS primitives are ray-conditioned and spatial-geometry-supervised, benefiting the map consistent and geometrically grounded under monocular camera condition. The chunk fusion is solved over the spatial geometry and poses via a global joint graph.}
    \label{fig:diagram}
\vspace{-5mm}
\end{figure*}

Our $\text{SG}^2$SLAM couples streaming spatial geometry with monocular-depth-aware 3DGS based on ray-conditioned kernels to deliver globally consistent mapping and tracking in monocular outdoor navigation. The frontend provides a spatial geometry prior in the form of poses, point maps, and per-pixel confidence maps in local level chunks, paired with a metric depth prior that supplies the metric scale correction. The backend optimizes a monocular-depth-aware 3DGS built on the frontend geometry prior for mapping and chunk fusion, where a ray-conditioned 3DGS kernel keeps the map consistent and aware as the viewing direction changes. Bundle adjustment combines a photometric loss with metric depth supervision, constraining scene scale and a point-consistency term, thus constraining pose drift. Across chunks, a joint Sim(3) optimization solves all chunk fusions in the global pose graph to mitigate the scale-induced drift accumulation that monocular SLAM suffers from along a sequence. The system is shown in Fig. \ref{fig:diagram}.

\subsection{Streaming Spatial Geometry Prior}
\label{ss:prior}

The frontend integrates a frozen reconstruction network $\Phi_\Theta$ \cite{han2026abot}. The network consumes a sliding window $\mathcal{V}_t$ of $N_w$ consecutive frames, and for every frame $i$ in the window, it predicts a local camera pose $\hat{\mathbf{T}}^{\mathcal{A}}_i$, a dense point map $\hat{\mathbf{X}}^{\mathcal{C}}_i$, and a per-pixel confidence $\hat{c}_i(\mathbf{u})$, where each point is predicted on the normalized ray $\tilde{\mathbf{u}}$ through the pixel $\mathbf{u}$, with pixel coordinates $(u_x,u_y)$ and pinhole intrinsics $(f_x,f_y,c_x,c_y)$, and with the range $\hat{z}_i(\mathbf{u})$ factored in log space.
\begin{equation}
\small
\begin{aligned}
\tilde{\mathbf{u}} &= \Big[\tfrac{u_x-c_x}{f_x},\ \tfrac{u_y-c_y}{f_y},\ 1\Big]^{\top},\\
\hat{\mathbf{X}}^{\mathcal{C}}_i(\mathbf{u}) &= \hat{z}_i(\mathbf{u})\,\tilde{\mathbf{u}}, \quad \hat{z}_i(\mathbf{u}) = \exp\!\big(\min(\ell_i(\mathbf{u}),\,\ell_{\max})\big),
\end{aligned}
\label{eq:ray}
\vspace{1mm}
\end{equation}
where $\ell_i$ is the predicted log-range and $\ell_{\max}$ caps the range of a single point. Consecutive frames are linked by the predicted relative pose $\hat{\boldsymbol{\Delta}}_i$,
\begin{equation}
\hat{\boldsymbol{\Delta}}_i = \big(\hat{\mathbf{T}}^{\mathcal{A}}_{i+1}\big)^{-1}
\hat{\mathbf{T}}^{\mathcal{A}}_i.
\label{eq:relative}
\end{equation}

\subsubsection{Chunking Setup}

A sequence of $N_{\mathrm{fr}}$ input frames is covered by chunks of $N_c$ frames overlapping by $N_o$ at a stride of $S_o = N_c - N_o$, so sequences of any length receive identical coverage.
\begin{equation}
\small
\mathcal{F}_k = \Big[\,kS_o,\ \min\big((k{+}1)N_c,\ N_{\mathrm{fr}}\big)\Big],
\label{eq:chunk}
\end{equation}
where $\mathcal{F}_k$ is the frame interval of chunk $k$. Each chunk is streamed by a forward call to $\Phi_\Theta$, with attention restarting from the first frame of the chunk.

\subsubsection{Metric Scale Correction}

Meanwhile, a metric depth network $\Psi$ \cite{piccinelli2024unidepth} is used to obtain the range $d^{\Psi}_i(\mathbf{u})$ and the validity mask $\mathrm{mask}^{\Psi}_i$ for correcting the chunk's local scale, where the depth ratio $r(\mathbf{u})$ from two networks is the factor for scaling the chunk. Over the pixels where both networks are valid,
{\small
\begin{align}
\mathcal{V} &= \{\mathbf{u}:\ \hat{z}_i(\mathbf{u})>0, d^{\Psi}_i(\mathbf{u})>0, \mathbf{u}\in\mathrm{mask}^{\Psi}_i \}, \nonumber \\
r(\mathbf{u}) &= \frac{d^{\Psi}_i(\mathbf{u})}{\hat{z}_i(\mathbf{u})}, \nonumber \\
\mathcal{I} &= \Big\{\mathbf{u}\in\mathcal{V}:\ \big|r(\mathbf{u})-\mathrm{med}(r)\big| < 3\cdot 1.4826\cdot\mathrm{MAD}(r)\Big\}, \nonumber \\
\bar{s}_k &= \mathrm{med}_{\mathbf{u}\in\mathcal{I}}\ r(\mathbf{u}) \label{eq:madin}
\end{align}
} \vspace{-3mm} \\ where $\mathcal{V}$ collects the pixels carrying a positive finite range from the prior, and $\mathcal{I}$ keeps the pixels with a ratio within a robust spread of the median. $1.4826\cdot\mathrm{MAD}(\cdot)$ converts the median absolute deviation (MAD) into the standard deviation of a Gaussian. An anchor scale $\bar{s}_k$ is obtained per chunk in the first frame of the chunk because every pose within the chunk is expressed against the first frame, and the same factor holds across the whole chunk.

The scaling is then applied by conjugation with a uniform dilation $\mathbf{S}_k = \mathrm{diag}(\bar{s}_k\mathbf{I}_3, 1)$. The scaled relative pose $\hat{\boldsymbol{\Delta}}^{(m)}_i$, the absolute pose
$\mathbf{T}^{\mathcal{A},(m)}_i$, and the point cloud $\mathbf{X}^{(m)}$ are marked throughout by the
superscript $(m)$.
\begin{equation}
\begin{aligned}
\hat{\boldsymbol{\Delta}}^{(m)}_i &= \mathbf{S}_k\,\hat{\boldsymbol{\Delta}}_i\,\mathbf{S}_k^{-1},\\
\mathbf{T}^{\mathcal{A},(m)}_{i+1} &= \mathbf{T}^{\mathcal{A},(m)}_i
   \big(\hat{\boldsymbol{\Delta}}^{(m)}_i\big)^{-1},\quad
   \mathbf{T}^{\mathcal{A},(m)}_0 = \mathbf{I}_4,\\
\mathbf{X}^{(m)} &= \bar{s}_k\,\hat{\mathbf{X}} .
\end{aligned}
\label{eq:applyscale}
\end{equation}

Thus, the frontend provides a metric reconstruction for every chunk: the poses $\{\mathbf{T}^{\mathcal{A},(m)}_{k,i}\}$, the point clouds
$\{\mathbf{X}^{(m)}_{k,i}\}$ with the confidences $\hat{c}_{k,i}$, and the
scaling anchor $\bar{s}_k$.

\subsection{Monocular Depth-Aware 3DGS}
\label{ss:backend}

To ensure that the scene is aware of and consistent with the view change under monocular camera conditions, we represent the scene with an extended Gaussian primitive:
\vspace{2mm}
\begin{equation}
\mathcal{G}_i = \left\{
\begin{aligned}
&\boldsymbol{\mu}_i \in \mathbb{R}^{6},\quad
 \boldsymbol{\Sigma}_i \in \mathbb{S}^{6},\quad o_i \in (0,1),\\
&\boldsymbol{\zeta}_i \in (0,\infty)^{3},\quad
 \mathbf{c}_i \in \mathbb{R}^{3}
\end{aligned}
\right\},
\label{eq:prim}
\vspace{1mm}
\end{equation}
whose mean $\boldsymbol{\mu}_i$ and covariance $\boldsymbol{\Sigma}_i$ contain spatial dimensions and a conditioning viewing-direction subspace, with $o_i$ as the base opacity and $\mathbf{c}_i$ as the base color. The joint covariance is parameterized by a block lower-triangular Cholesky factor
$\boldsymbol{\Sigma}_i=\mathbf{L}_i\mathbf{L}_i^{\top}$ with
\begin{equation}
\mathbf{L}_i = \begin{bmatrix}
\mathbf{R}_i\,\mathrm{diag}(\mathbf{s}_{i,1:3}) & \mathbf{0}\\[2pt]
\mathbf{L}_{i,21} & \mathbf{L}_{i,22}
\end{bmatrix},
\quad
\mathbf{R}_i = \mathbf{I_{3}} + [\mathbf{a}_i]_{\times},
\end{equation}
where the spatial block combines a first-order rotation $\mathbf{R}_i$ obtained from the skew-symmetric map $\mathbf{a}_i$ with anisotropic scales $\mathbf{s}_{i}$, and the free sub-diagonal blocks couple the spatial and conditioning subspaces so that the primitive may encode view-dependent geometry. $o_i$ is the base opacity, while the colors $\mathbf{c}_i$ are stored directly as base RGB, and the per-dimension shape parameter is activated as $\boldsymbol{\zeta}_i = 4\exp(\tilde{\boldsymbol{\zeta}}_i)$, where $\tilde{\boldsymbol{\zeta}_i} = \min(\boldsymbol{\zeta}_i/ 4, 1)$ are the damped exponents controlling how strongly the primitive conditions affect the ray.

\subsubsection{Ray-Conditioned Kernel}

At rendering time, each primitive is queried with the unit viewing ray from the camera center $\mathbf{o}_{\mathrm{cam}}$.
\vspace{-1mm}
\begin{equation}
\mathbf{q}_i = \frac{\boldsymbol{\mu}^{(1)}_i - \mathbf{o}_{\mathrm{cam}}}
{\big\|\boldsymbol{\mu}^{(1)}_i - \mathbf{o}_{\mathrm{cam}}\big\|_2},
\label{eq:query}
\end{equation} \\ where $\boldsymbol{\mu}^{(1)}_i$ and $\boldsymbol{\mu}^{(2)}_i$ denote the spatial and conditioning blocks of $\boldsymbol{\mu}_i$. Partitioning the joint covariance with conformal blocks $\mathbf{V}$.
\vspace{3mm}
\begin{equation}
    \boldsymbol{\Sigma}_i = \begin{bmatrix}\mathbf{V}_{11}&\mathbf{V}_{12}\\
\mathbf{V}_{21}&\mathbf{V}_{22}\end{bmatrix},
\end{equation}
the primitive responds to the query through the regression coefficient
$\mathbf{A}_i=\mathbf{V}_{12}\mathbf{V}_{22}^{-1}$ damped by
$\bar{\boldsymbol{\zeta}}_i=\min(\boldsymbol{\zeta}_i/4,\mathbf{1})$, yielding the conditional mean and covariance:
\begin{align}
\boldsymbol{\mu}_i(\mathbf{q}) &= \boldsymbol{\mu}^{(1)}_i
   + \mathbf{A}_i\,\mathrm{diag}(\bar{\boldsymbol{\zeta}}_i)\,
     \big(\mathbf{q}-\boldsymbol{\mu}^{(2)}_i\big),
\label{eq:condmean}\\[2pt]
\boldsymbol{\Sigma}_i(\mathbf{q}) &= \mathbf{V}_{11}
   - \mathbf{A}_i\,\mathrm{diag}(\bar{\boldsymbol{\zeta}}_i)\,\mathbf{V}_{21}.
\label{eq:condcov}
\end{align}

Equations~\eqref{eq:condmean} and \eqref{eq:condcov} are the Schur-complement conditional statistics of the joint Gaussian, with the regression term attenuated by $\bar{\boldsymbol{\zeta}}_i$ so that the primitive may interpolate between view-independent behavior ($\boldsymbol{\zeta}\!\to\!\mathbf{0}$) and fully conditioned behavior. In parallel, the base opacity is modulated by the whitened deviation of the actual ray $\mathbf{y}_i$ from the conditioning mean,
\begin{equation}
\small
o_i(\mathbf{q}) = o_i \prod_{k=1}^{3}
\Big(1-\tanh\big(\mathbf{y}_{i,k}^{2}\big)\Big)^{\zeta_{i,k}}, \ 
\mathbf{y}_i = \mathbf{L}_{i,22}^{-1}\big(\mathbf{q}-\boldsymbol{\mu}^{(2)}_i\big),
\label{eq:opacitygate}
\end{equation}
which acts as a smooth, compactly supported gate: a primitive remains opaque for viewing rays consistent with its conditioning subspace and vanishes smoothly otherwise. Because the conditioning subspace is populated and supervised by the monocular spatial geometry prior above and the depth objective below, we refer to the resulting kernel as \emph{monocular depth-aware}.

\subsubsection{Rasterization}

The conditional mean and covariance are projected onto the image plane by the standard exponential moving average transform, yielding a 2D conic and the Mahalanobis distance $\sigma_i(\mathbf{u})=\tfrac{1}{2}\boldsymbol{\delta}^{\top}\boldsymbol{\Sigma}^{-1}_{i,2D}, \ \boldsymbol{\delta}\in[0,1)$ of pixel $\mathbf{u}$ from the projected center. Whereas classical 3DGS evaluates the Gaussian kernel $\alpha_i = o_i\,e^{-\sigma_i}$, we employ a ray-conditioned kernel.
\begin{equation}
\alpha_i(\mathbf{u}) = \min\!\Big(\alpha_{\max},\
o_i(\mathbf{q})\,\big(1-\sigma_i(\mathbf{u})\big)^{\zeta_i}\Big),
\label{eq:zetakernel}
\end{equation}
which replaces the unbounded Gaussian tails with a polynomially decaying profile that vanishes at $\sigma_i=1$, eliminating inter-primitive blending artifacts at grazing viewing angles while retaining a closed-form derivative with respect to $\zeta_i$. Color and depth are composited front-to-back along each ray by alpha-blending.
\begin{equation}
\small
\hat{\mathbf{C}}(\mathbf{u}) = \sum_{i\in\mathcal{N}(\mathbf{u})} \mathbf{c}_i\,
\alpha_i \prod_{j<i}(1-\alpha_j),
\end{equation}

\begin{equation}
\small
\hat{D}(\mathbf{u}) = \sum_{i\in\mathcal{N}(\mathbf{u})} \bar{z}_i\,
\alpha_i \prod_{j<i}(1-\alpha_j),
\label{eq:compositing}
\end{equation}
where $\mathcal{N}(\mathbf{u})$ is the depth-sorted primitive set hitting pixel $\mathbf{u}$, and $\bar{z}_i$ is the camera-frame depth of $\boldsymbol{\mu}_i(\mathbf{q})$. The expected depth $\hat{D}$ is accumulated using the same weights as color and is fully differentiable through the kernel of Eq.~\eqref{eq:zetakernel}. 

\subsubsection{Map Optimization} 

Keyframes are selected within each chunk by scoring every frame based on the number of finite and confident points above the threshold. When a keyframe arrives, its metrically anchored point map is appended to the map, where an isotropic scale is initialized by the mean distance to its three nearest neighbors.
\begin{equation}
\mathbf{s}_i = \mathrm{mean}_{j\in\mathrm{NN}_3(i)}\
\big\|\hat{\mathbf{X}}^{\mathcal{C}}(\mathbf{u}_i)-\hat{\mathbf{X}}^{\mathcal{C}}(\mathbf{u}_j)\big\|_2,
\label{eq:knninit}
\end{equation}
so that primitive footprints match the local sampling density of the prior. Mapping then proceeds as windowed bundle adjustment over the sliding set of keyframes, jointly optimizing the primitive parameters and the keyframe poses $\mathbf{T}_k$. The objective combines a photometric term against input images $\mathbf{I}$, monocular metric depth supervision, and geometric regularization edges with their corresponding weighting coefficients $\lambda$.
\begin{align}
\mathcal{L}_{\mathrm{map}} &= \mathcal{L}_{\mathrm{ph}}
 + \lambda_{d}\mathcal{L}_{\mathrm{dep}} + \lambda_{pc}\mathcal{L}_{pc}
 + \lambda_{r}\mathcal{L}_{rel},
\label{eq:maploss}\\[2pt]
\mathcal{L}_{\mathrm{ph}}
  &= (1-\lambda_{\mathrm{ssim}})\big\lVert\hat{\mathbf{C}}-\mathbf{I}\big\rVert_1
   + \lambda_{\mathrm{ssim}}\big(1-\mathrm{SSIM}(\hat{\mathbf{C}},\mathbf{I})\big),
\label{eq:phloss}\\[2pt]
\mathcal{L}_{\mathrm{dep}}
  &= \frac{1}{|\Omega|}\sum_{\mathbf{u}\in\Omega}
     \varphi(\mathbf{u})\,\rho_{\delta_d}\!\Big(
     \hat{D}(\mathbf{u}) - d^{\Psi}_k(\mathbf{u})\Big),
\label{eq:depthloss}\\[2pt]
\mathcal{L}_{\mathrm{pc}}
  &= \frac{1}{|\mathcal{P}_k|}\sum_{\mathbf{u}\in\mathcal{P}_k}
\rho_{\delta_p}\!\Big(\big\|\mathbf{X}_k\big\|_2\Big),
\label{eq:pcloss}\\[2pt]
\mathcal{L}_{rel}
  &= \big\|\mathbf{R}_{E}-\mathbf{I}_3\big\rVert_F^2 + \big\lVert\mathbf{t}_{E}\big\rVert_2^2,
  \quad
  \mathbf{E} = \hat{\mathbf{T}}_{ij}\,\hat{\boldsymbol{\Delta}}_{ij}^{-1},
\label{eq:relloss}
\end{align}
where $\rho_\delta$ is the Huber function, and the mask $\Omega$ is the set of pixels that the prior supports with confident geometry, and $\varphi$ weights the depth residual by the validity intersection with the depth network's own confidence. In the valid point set $\mathcal{P}_k$, the pose and spatial geometry tie $\mathbf X_k(\mathbf u)=\hat{\mathbf{X}}^{\mathcal{W}}_k(\mathbf u)
-\mathbf T_k\hat{\mathbf{X}}^{\mathcal{C}}_k(\mathbf u)$
pins the keyframe's stored geometry to the world frame to prevent pose error from propagating into the built map, while the relative pose term $\mathcal{L}_{rel}$ penalizes the discrepancy $\mathbf{E}$ between the currently estimated inter-keyframe transform $\hat{\mathbf{T}}_{ij}=\hat{\mathbf{T}}_i^{-1}\hat{\mathbf{T}}_j$ and its prior value.

The mapping is cumulative, while a new chunk is optimized against both their input images and spatial geometry, refining a chunk's geometry without disturbing the built maps. The global joint trajectory and chunk fusion are described below.

\begin{figure*}[h]
    \centering
    \begin{subfigure}[b]{0.195\textwidth}
        \includegraphics[width=\textwidth]{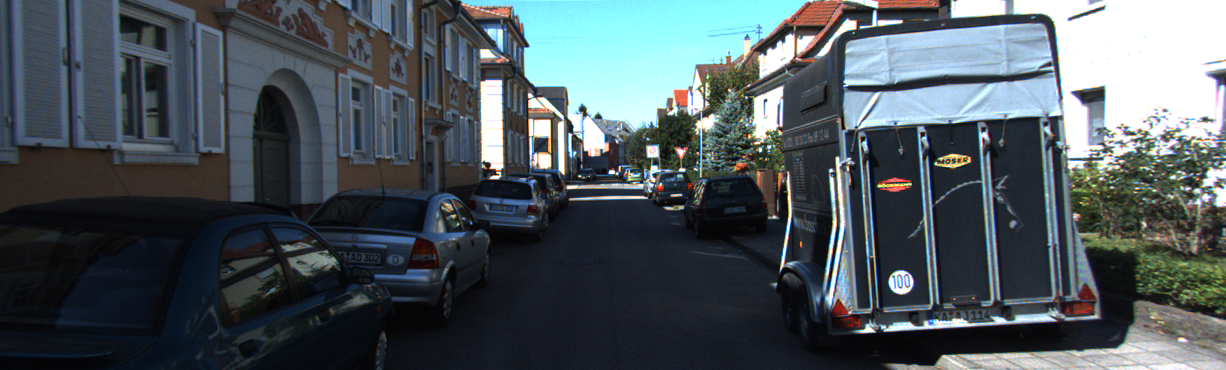}
    \end{subfigure}
    \hfill
    \begin{subfigure}[b]{0.195\textwidth}
        \includegraphics[width=\textwidth]{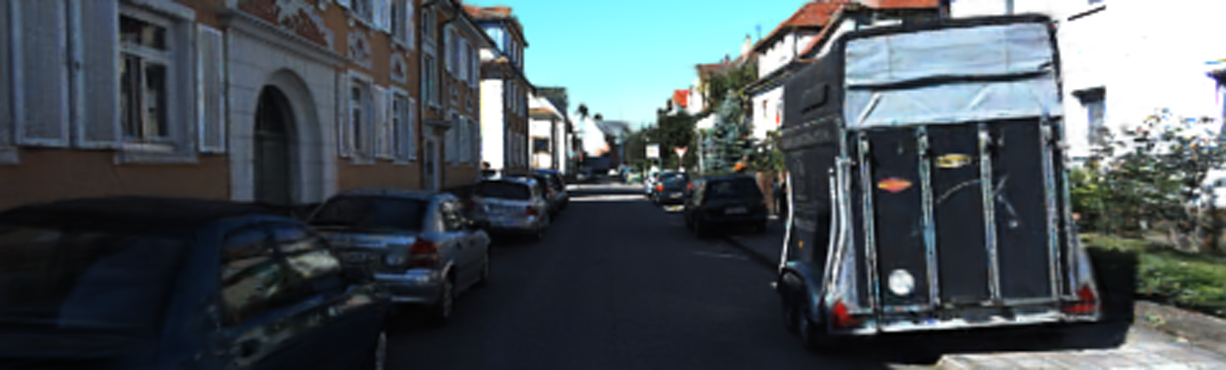}
    \end{subfigure}
    \hfill
    \begin{subfigure}[b]{0.195\textwidth}
        \includegraphics[width=\textwidth]{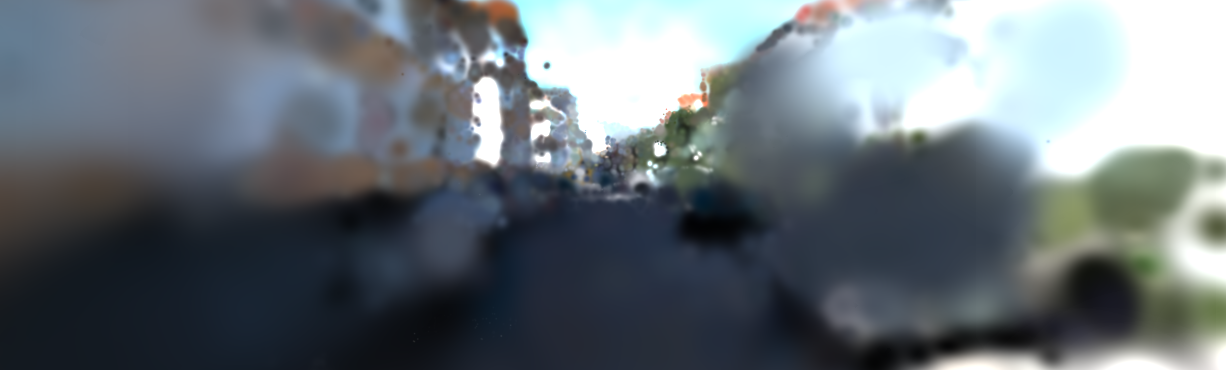}
    \end{subfigure}
    \hfill
    \begin{subfigure}[b]{0.195\textwidth}
        \includegraphics[width=\textwidth]{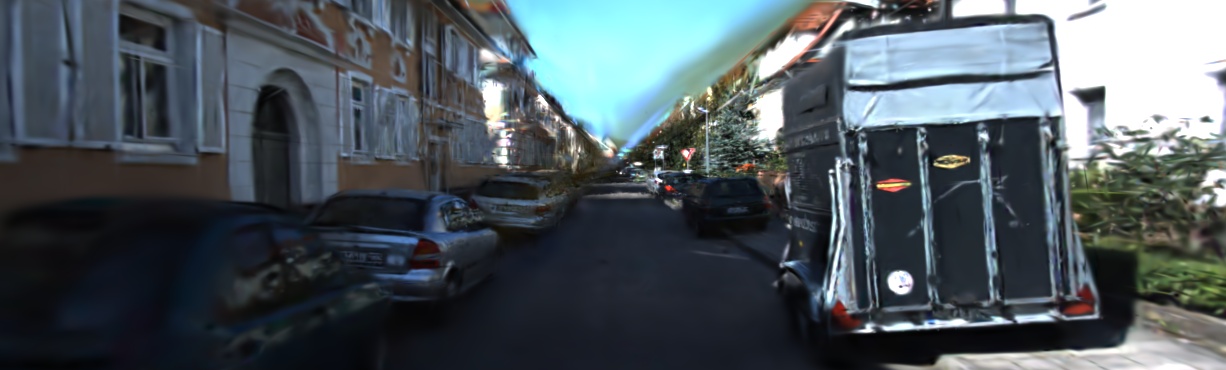}
    \end{subfigure}
    \hfill
    \begin{subfigure}[b]{0.195\textwidth}
        \includegraphics[width=\textwidth]{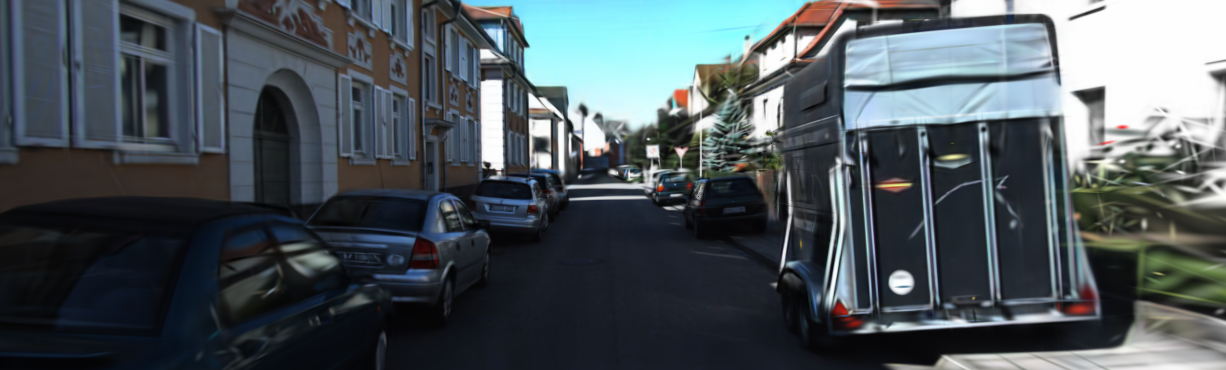}
    \end{subfigure}

    \vspace{1ex} 

    \begin{subfigure}[b]{0.195\textwidth}
        \includegraphics[width=\textwidth]{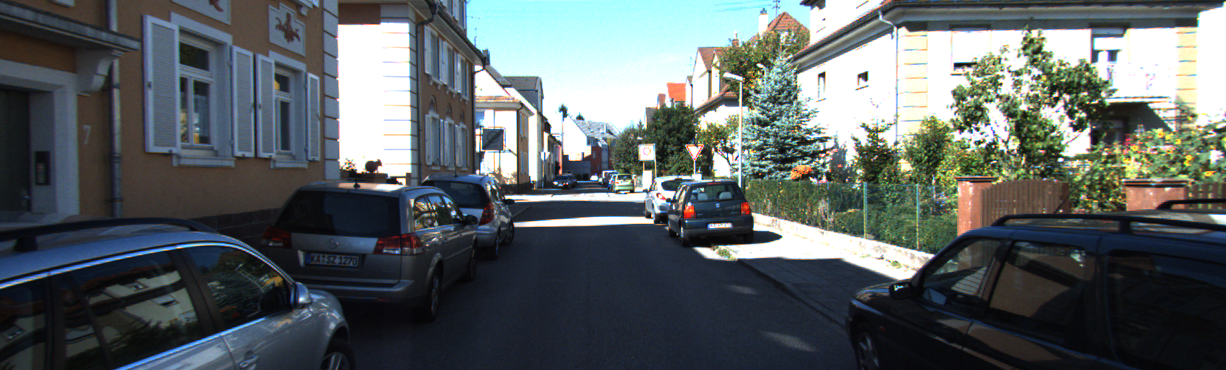}
    \end{subfigure}
    \hfill
    \begin{subfigure}[b]{0.195\textwidth}
        \includegraphics[width=\textwidth]{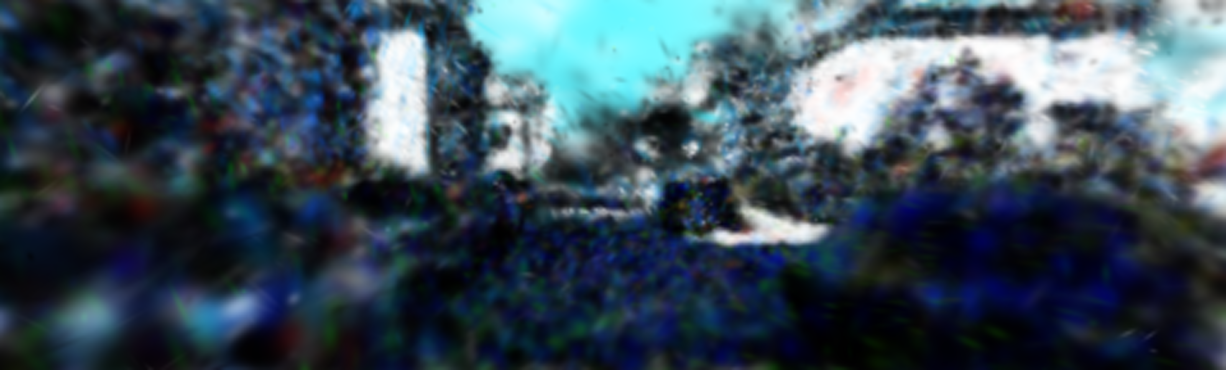}
    \end{subfigure}
    \hfill
    \begin{subfigure}[b]{0.195\textwidth}
        \includegraphics[width=\textwidth]{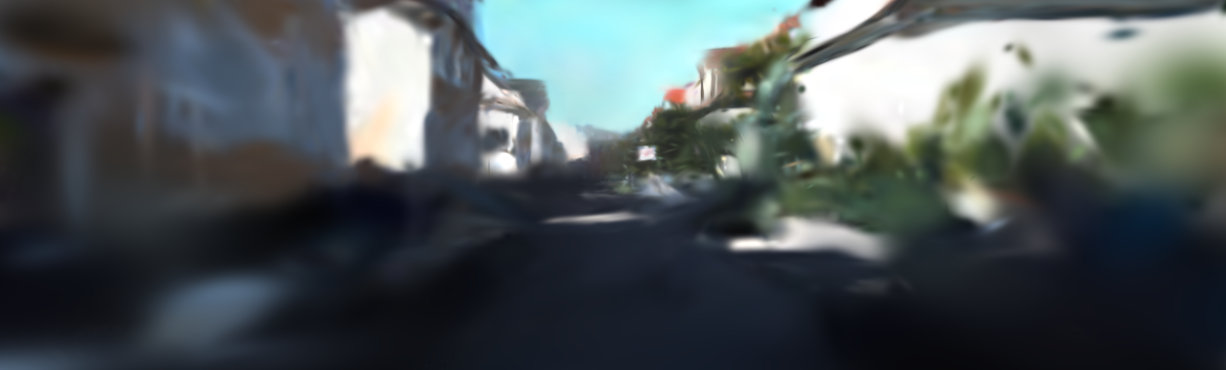}
    \end{subfigure}
    \hfill
    \begin{subfigure}[b]{0.195\textwidth}
        \includegraphics[width=\textwidth]{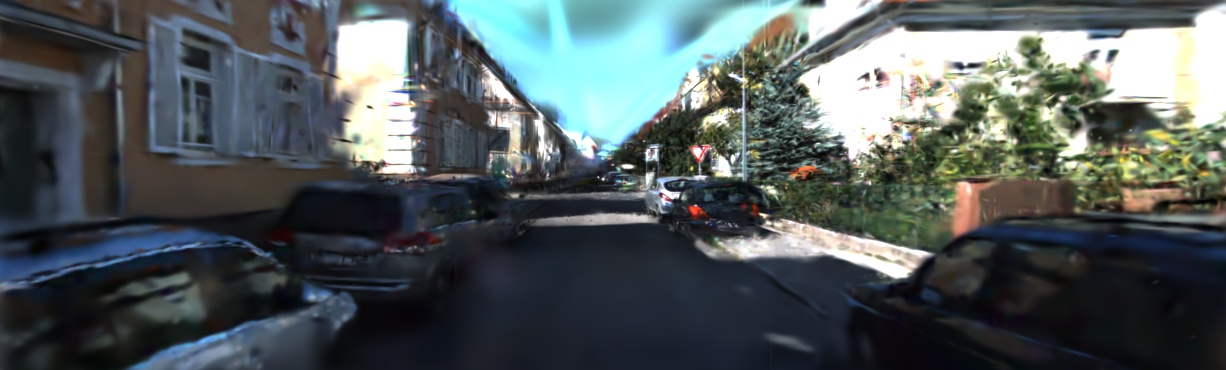}
    \end{subfigure}
    \hfill
    \begin{subfigure}[b]{0.195\textwidth}
        \includegraphics[width=\textwidth]{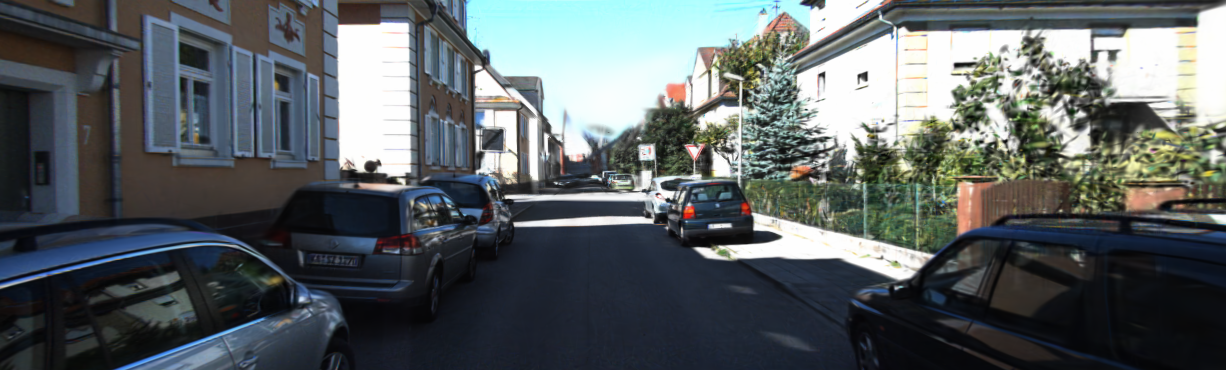}
    \end{subfigure}

    \vspace{1ex}

    \begin{subfigure}[b]{0.195\textwidth}
        \includegraphics[width=\textwidth]{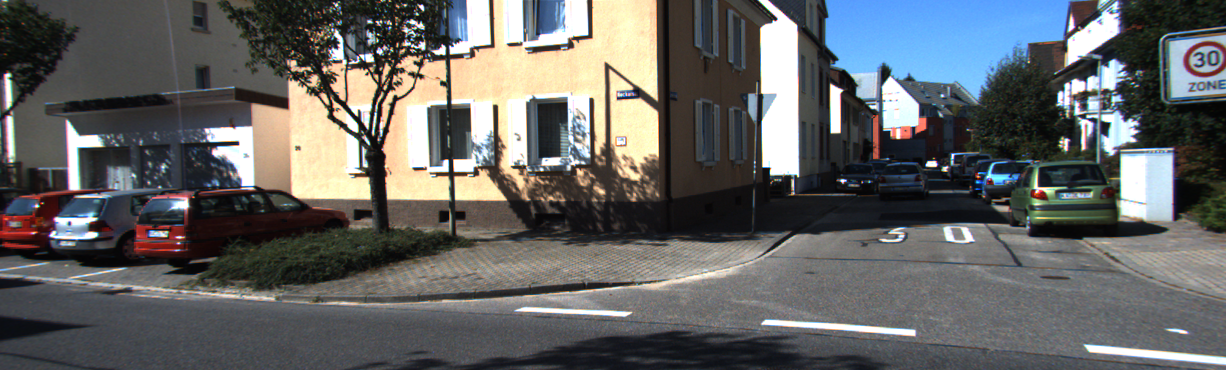}
        \caption{GT}
    \end{subfigure}
    \hfill
    \begin{subfigure}[b]{0.195\textwidth}
        \includegraphics[width=\textwidth]{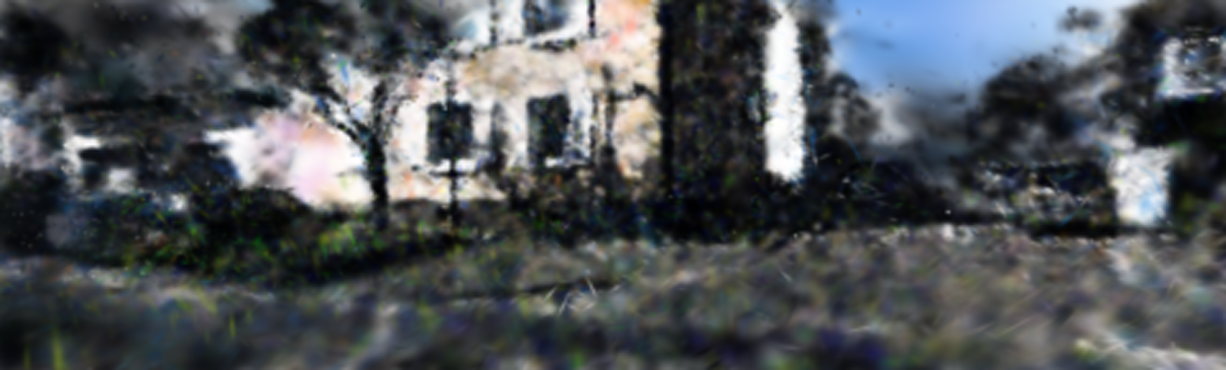}
        \caption{GigaSLAM}
    \end{subfigure}
    \hfill
    \begin{subfigure}[b]{0.195\textwidth}
        \includegraphics[width=\textwidth]{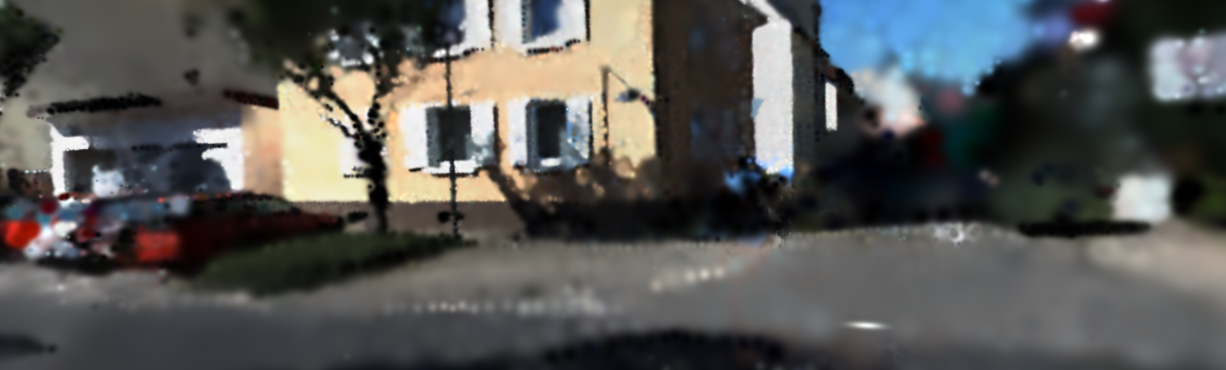}
        \caption{S3PO-GS}
    \end{subfigure}
    \hfill
    \begin{subfigure}[b]{0.195\textwidth}
        \includegraphics[width=\textwidth]{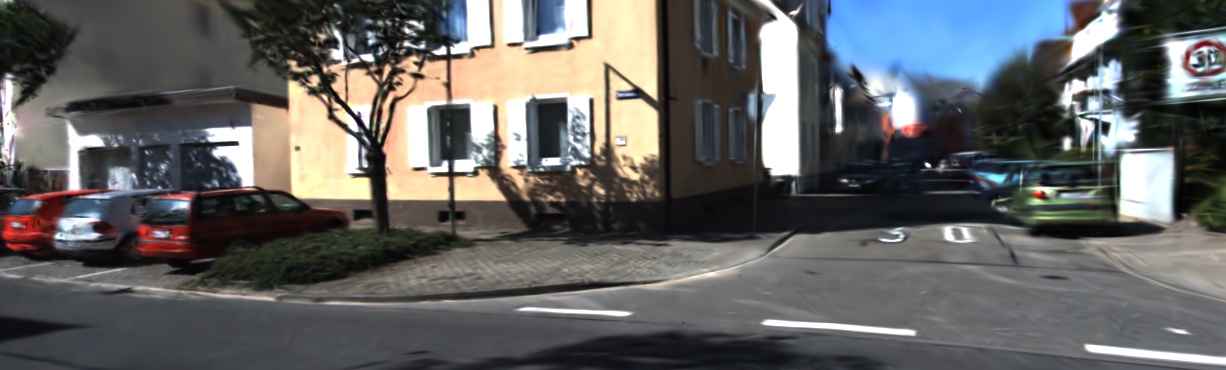}
        \caption{DiskChunGS}
    \end{subfigure}
    \hfill
    \begin{subfigure}[b]{0.195\textwidth}
        \includegraphics[width=\textwidth]{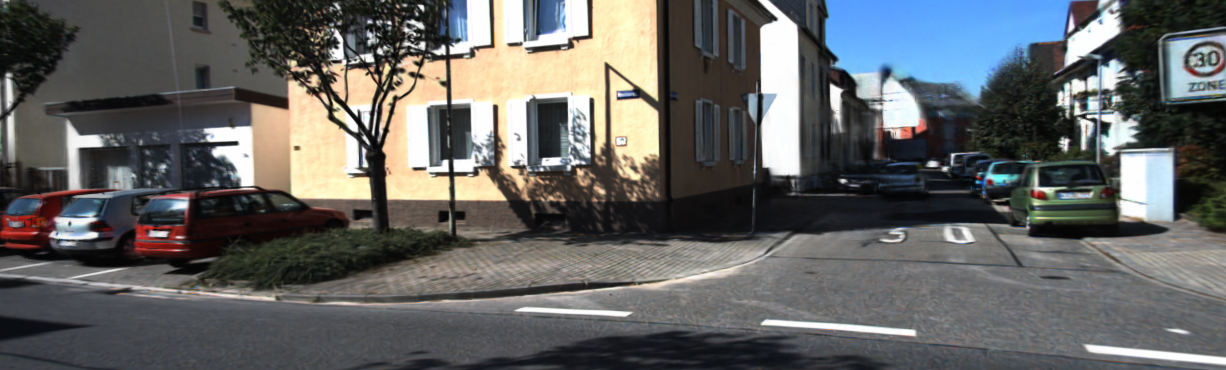}
        \caption{\textbf{Ours}}
    \end{subfigure}

    \vspace{-2mm}
    \caption{Qualitative results of 3DGS rendered RGB image on the KITTI dataset. Our $\text{SG}^2$SLAM renders better details both in foreground and background based on our monocular-depth-aware 3DGS kernel.}
    \label{fig:3x5_grid}
    \vspace{-6mm}
\end{figure*}

\subsection{Global Joint Trajectory and Chunk Fusion}
\label{ss:graph}

Every chunk produced by Sec.~\ref{ss:prior} and refined by Sec.~\ref{ss:backend} constitutes a self-consistent reconstruction in a local anchor frame, and we fuse every chunk beyond the initial chunk with a similarity to reach the global frame. Let chunk $k$ take the similarity
$\mathbf{G}_k\in Sim(3)$,
\begin{equation}
\begin{aligned}
\mathbf{X}^{(g)} &= \sigma_k\,\mathbf{R}_k\,\mathbf{X}^{(m)} + \mathbf{b}_k,\\
\mathbf{G}_k &= \begin{bmatrix}\sigma_k\,\mathbf{R}_k & \mathbf{b}_k\\
\mathbf{0}^{\top} & 1\end{bmatrix},\qquad \mathbf{G}_1 = \mathbf{I}_4,
\end{aligned}
\label{eq:placement}
\end{equation}
where $\sigma_k>0$ is the scale factor of the global placement similarity $\mathbf{G}_k$, and $\mathbf{R}_k\in SO(3)$, $\mathbf{b}_k\in\mathbb{R}^3$ collect the scale, rotation, and translation for placing chunk $k$, and the superscript $(g)$ marks the global frame. The constraint $\mathbf{G}_1=\mathbf{I}_4$ fixes the seven degrees of freedom shared by the whole fusion, leaving $7(K{-}1)$ free parameters for $K$ chunks to be optimized.

We use a closed-form similarity fit between two matched point sets, which recovers the scale by,
\begin{equation}
\label{eq:umeyama}
s^{\mathrm{fit}} = \frac{\sum_{n}\big\langle \mathbf{y}_n-\bar{\mathbf{y}},\ \mathbf{R}^{\star}(\mathbf{x}_n-\bar{\mathbf{x}})\big\rangle}{\sum_{n}\big\lVert \mathbf{x}_n-\bar{\mathbf{x}}\big\rVert_2^2},
\end{equation}
where $\mathbf{x}_n$ and $\mathbf{y}_n$ are the $n$-th matched points of the two sets, $\bar{\mathbf{x}}$ and $\bar{\mathbf{y}}$ are the two centroids, and $\mathbf{R}^{\star}$ is the rotation that aligns the sets. Applied to two point clouds, Eq.~\eqref{eq:umeyama} therefore returns a scale that carries the density ratio of the two matching spatial regions.

\subsubsection{Chain Constraints}

Adjacent chunks $A$ and $B$ share the $N_o$ frames of the overlap defined in Eq.~\eqref{eq:chunk}, where the camera centers of the shared frames constrain the relative similarity. Let $s^{\mathrm{ov}}_{AB}$ denote the scale that Eq.~\eqref{eq:umeyama} returns on the shared camera centers of the two chunks, and let $\bar{s}_A$ and $\bar{s}_B$ denote the metric anchors of the two chunks from Eq.~\eqref{eq:madin}. The edge scale combines
the two using the geometric mean,
\begin{equation}
\label{eq:chainscale}
\hat{\sigma}_{AB} = \sqrt{s^{\mathrm{ov}}_{AB}\cdot \frac{\bar{s}_A}{\bar{s}_B}}.
\end{equation}

Camera centers are free from the influence of sampling density because the spatial geometry prior reports a single center per frame, which leaves the chain scale independent of the density bias of Eq.~\eqref{eq:umeyama}.

Rotation and translation are recovered by the spatial geometry across the shared window. Let $\mathbf{c}^{\mathrm{cam}}_{k,i}$ denote the camera center of frame $i$ as placed by chunk $k$, let $\mathcal{O}_{AB}=\mathcal{F}_A\cap\mathcal{F}_B$ denote the shared frame set, and write
\begin{equation}
\label{eq:centres}
\bar{\mathbf{c}}_k=\frac{1}{|\mathcal{O}_{AB}|}\sum_{i\in\mathcal{O}_{AB}}\mathbf{c}^{\mathrm{cam}}_{k,i},
\quad
\tilde{\mathbf{c}}_{k,i}=\mathbf{c}^{\mathrm{cam}}_{k,i}-\bar{\mathbf{c}}_k ,
\end{equation}
the centroid over the shared frames and the calibrated center of frame $i$. The edge $(A,B)$ carries the similarity mapping chunk $B$ into the frame of chunk $A$, so the seed aligns the calibrated centers of $B$ with the calibrated centers of $A$.
\begin{equation}
\begin{aligned}
\mathbf{R}^{(0)}_{AB} &= \arg\max_{\mathbf{R}\in SO(3)}\ 
   \sum_{i\in\mathcal{O}_{AB}}
   \big\langle \tilde{\mathbf{c}}_{A,i},\ \mathbf{R}\,\tilde{\mathbf{c}}_{B,i}\big\rangle,\\
\mathbf{b}^{(0)}_{AB} &= \bar{\mathbf{c}}_A - \mathbf{R}^{(0)}_{AB}\,\bar{\mathbf{c}}_B ,
\end{aligned}
\label{eq:chainrt}
\end{equation}
where the superscript $(0)$ marks a seed later refined by Eq.~\eqref{eq:chainicp}.

The refinement aims to align the point clouds of the shared frames. With $\mathcal{C}$ the trimmed correspondences between the two clouds, the refinement solves
\begin{equation}
\small
\label{eq:chainicp}
(\mathbf{R}_{AB},\mathbf{b}_{AB})=\arg\min_{\mathbf{R},\mathbf{b}}\ 
\sum_{(\mathbf{x}_A,\mathbf{x}_B)\in\mathcal{C}}
\big\|\hat{\sigma}_{AB}\,\mathbf{R}\,\mathbf{x}_A+\mathbf{b}-\mathbf{x}_B\big\|_2^2 ,
\end{equation}
over points $\mathbf{x}_A$ of cloud $A$ and $\mathbf{x}_B$ of cloud $B$, with the scale of each correspondence pinned to $\hat{\sigma}_{AB}$. Pinning the scale leaves the refinement free to adjust rotation and translation alone.

\begin{figure}[h]
    \centering
    \begin{subfigure}[b]{0.156\textwidth}
        \includegraphics[width=\textwidth]{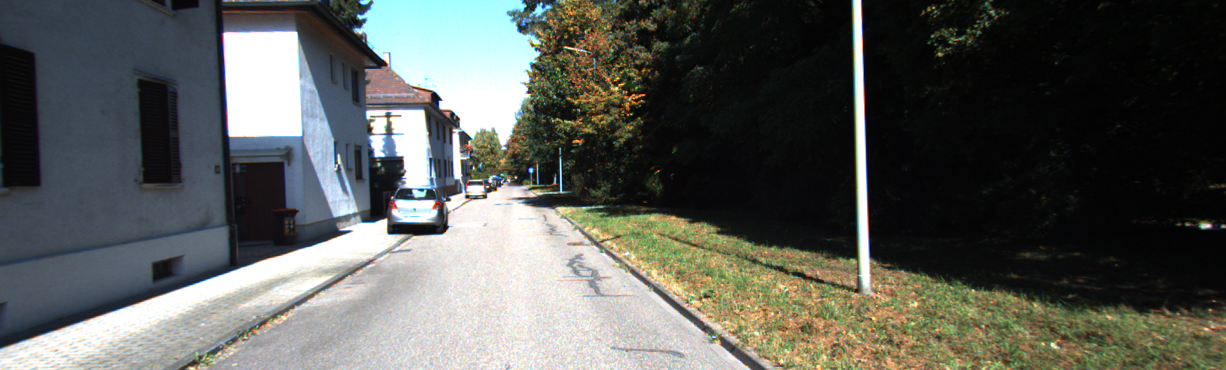}
    \end{subfigure}
    \hfill
    \begin{subfigure}[b]{0.156\textwidth}
        \includegraphics[width=\textwidth]{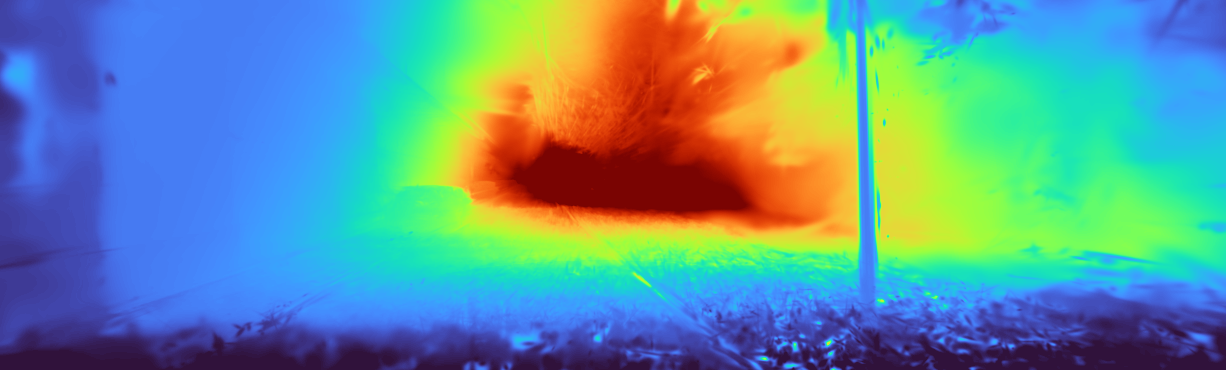}
    \end{subfigure}
    \hfill
    \begin{subfigure}[b]{0.156\textwidth}
        \includegraphics[width=\textwidth]{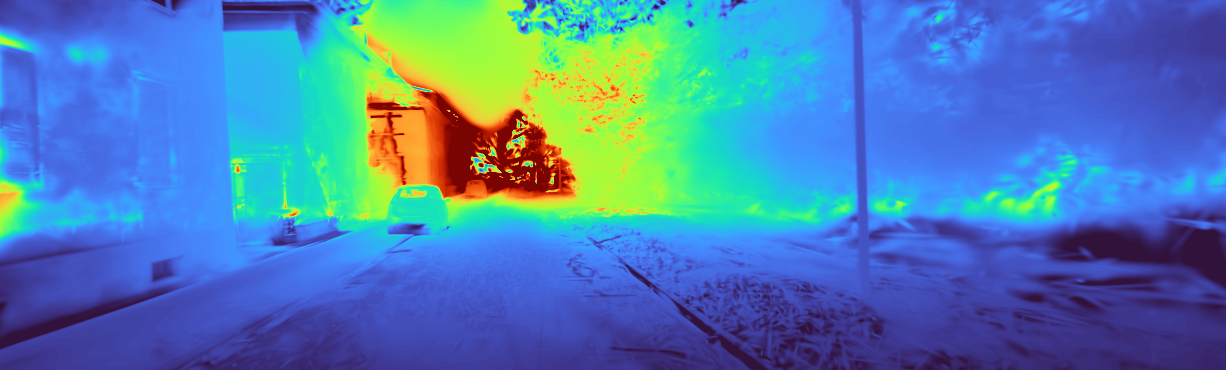}
    \end{subfigure}

    \vspace{1ex} 

    \begin{subfigure}[b]{0.156\textwidth}
        \includegraphics[width=\textwidth]{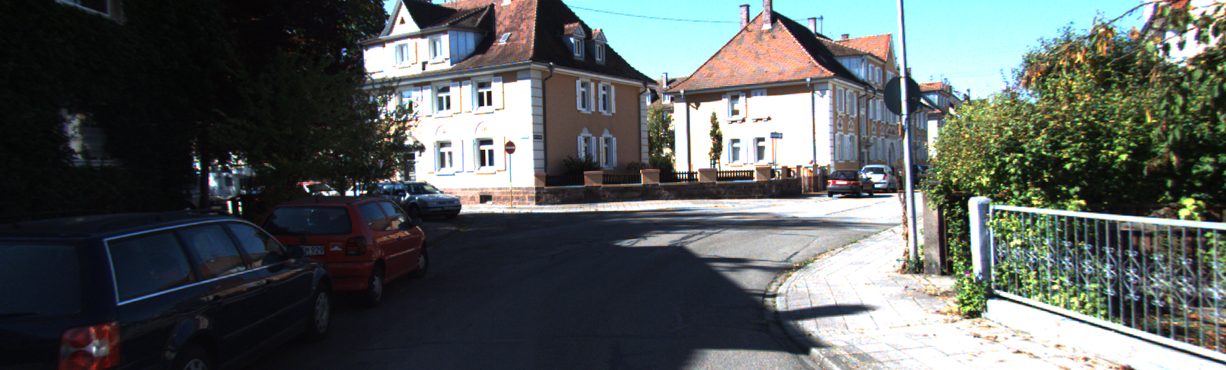}
    \end{subfigure}
    \hfill
    \begin{subfigure}[b]{0.156\textwidth}
        \includegraphics[width=\textwidth]{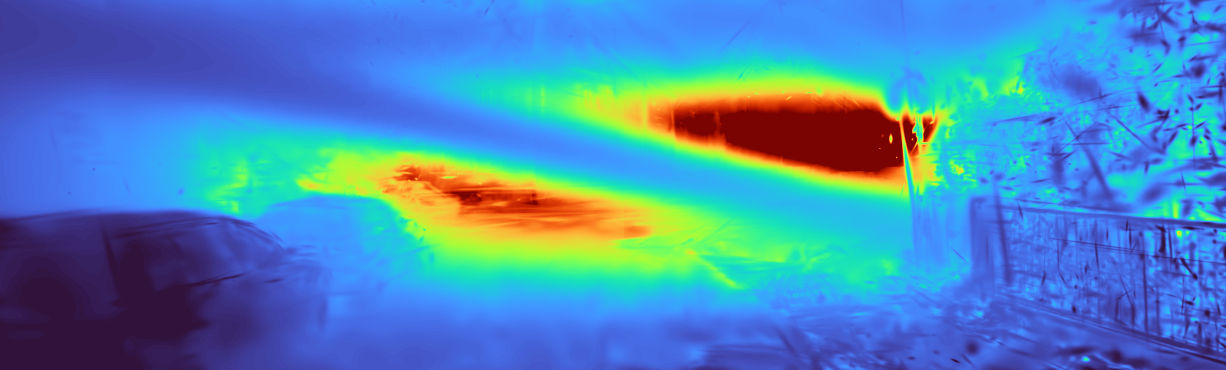}
    \end{subfigure}
    \hfill
    \begin{subfigure}[b]{0.156\textwidth}
        \includegraphics[width=\textwidth]{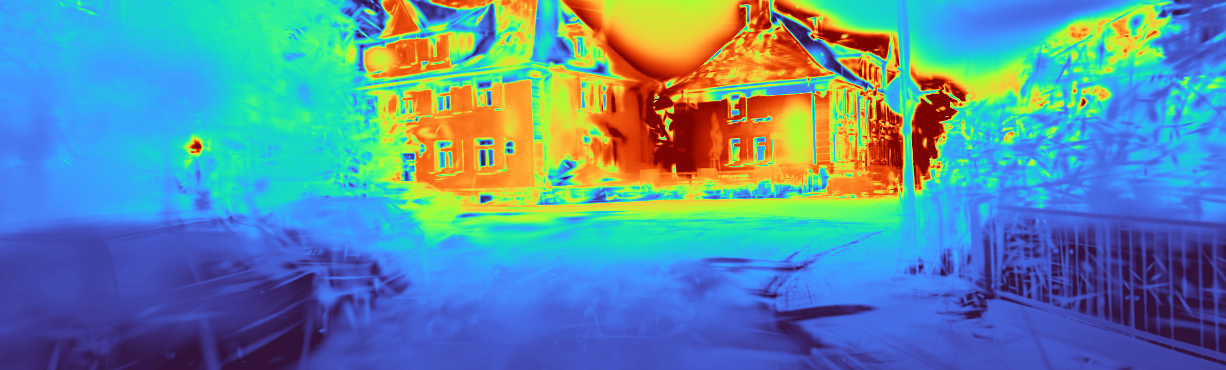}
    \end{subfigure}

    \vspace{1ex}

    \begin{subfigure}[b]{0.156\textwidth}
        \includegraphics[width=\textwidth]{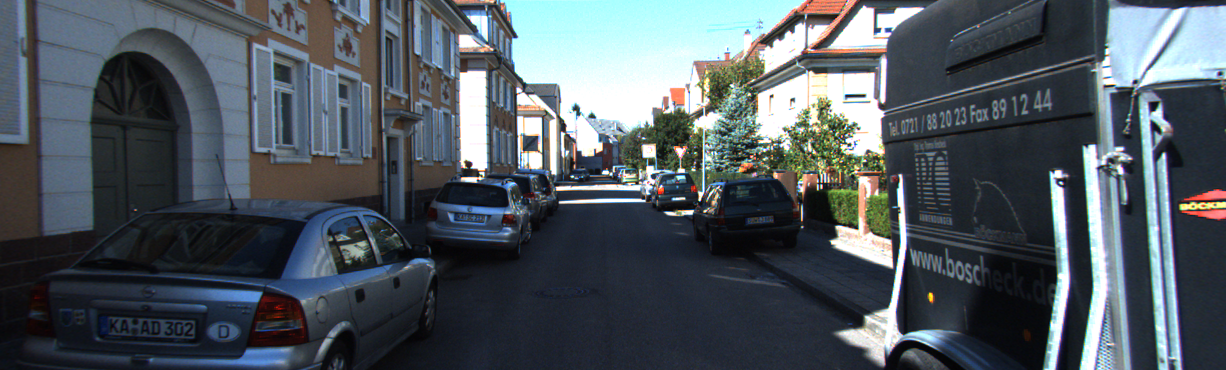}
        \caption{GT}
    \end{subfigure}
    \hfill
    \begin{subfigure}[b]{0.156\textwidth}
        \includegraphics[width=\textwidth]{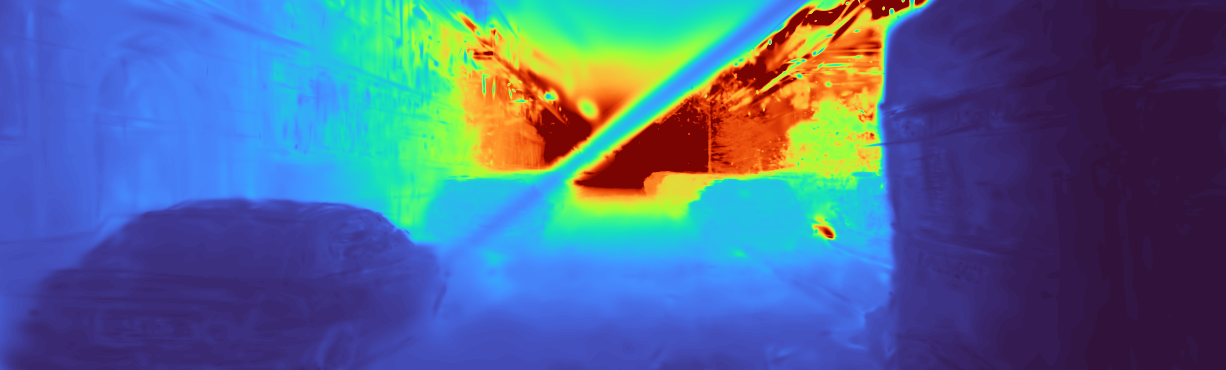}
        \caption{DiskChunGS}
    \end{subfigure}
    \hfill
    \begin{subfigure}[b]{0.156\textwidth}
        \includegraphics[width=\textwidth]{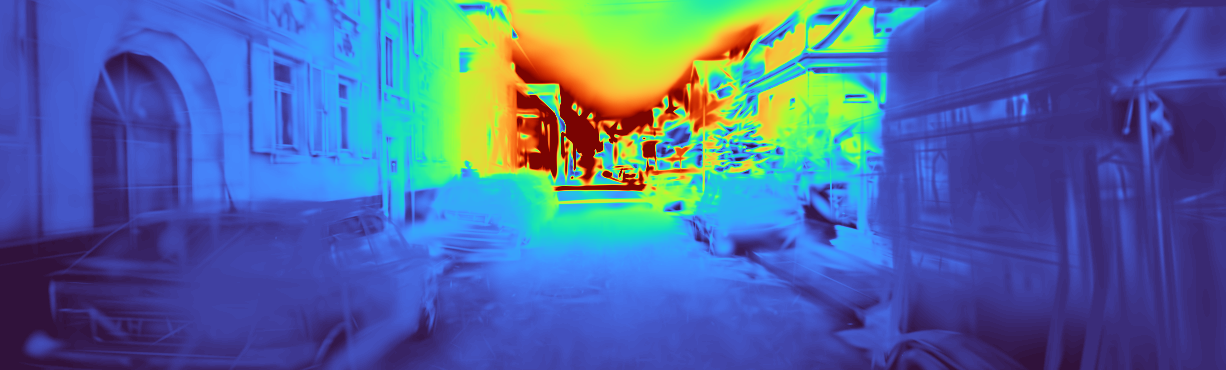}
        \caption{\textbf{Ours}}
    \end{subfigure}

    \vspace{-2mm}
    \caption{Qualitative results of 3DGS rendered depth image on the KITTI dataset. Our rendered depth effectively removes the floating primitives that are harmful to the downstream navigation.}
    \label{fig:3x3_grid}
    \vspace{-3mm}
\end{figure}

\subsubsection{Loop Constraints}

A revisit between non-adjacent chunks closes a cycle in the pose graph that a chain leaves open, which is what makes the drift accumulated between the two observations observable and correctable.

Let $d^{A}_{a_n}(p)$ denote the range predicted for pixel $p$ of frame $a_n$ in chunk $A$, and $d^{B}_{b_n}(p)$ denote the same quantity in chunk $B$. Let $\mathcal{Q}_n$ hold the pixels where both ranges lie in the band that keeps the sky and far field out, where a predicted range is poorly conditioned. The edge scale takes the median of the per-pixel ratios inside a match and combines the matches by retrieval score.
\vspace{-2mm}
\begin{equation}
\small
\begin{aligned}
\rho_n &= \mathrm{med}_{p\in\mathcal{Q}_n}\
\frac{d^{A}_{a_n}(p)}{d^{B}_{b_n}(p)},\\[2pt]
\hat{\sigma}_{AB} &= \frac{\sum_{n}\gamma_n\,\rho_n}{\sum_{n}\gamma_n},
\quad \gamma_n = c_n^{2}.
\end{aligned}
\label{eq:loopscale}
\end{equation}
with each pair $(a_n,b_n)$ scored by the normalized cross correlation $c_n\in(0,1]$ between the two thumbnail descriptors. The median runs over the pixels observing a single patch, which removes the density bias skewing cloud-to-cloud alignment: the patch is sampled densely from every direction at once, so the two clouds contribute comparable numbers of pixels to every region, leaving the ratio free of density bias.

Rotation follows the same correspondences, with each pairing two views of a single physical camera. Let $\mathbf{R}_{A,a_n}$ and $\mathbf{R}_{B,b_n}$ denote the orientations of the two frames; then the consensus rotation is defined as:
\begin{equation}
\hat{\mathbf{R}}_{AB} = \Big(\textstyle\sum_{n=1}^{N_m}\nu_n\,
\mathbf{R}_{A,a_n}\mathbf{R}_{B,b_n}^{\top}\Big)^{\dagger}, \quad \nu_n= c_n^3>0,
\label{eq:looprot}
\end{equation}
where $\nu_n$ weights a match in the consensus, and $(\cdot)^{\dagger}$ denotes the orthogonal factor and keeps the output in $SO(3)$. Eq.~\eqref{eq:umeyama} applied to a set of near-coincident centers admits no scale to recover, so the rotation follows from the orientation consensus rather than from a center fit.

\begin{figure}[h]
    \centering
    \includegraphics[width=0.9\linewidth]{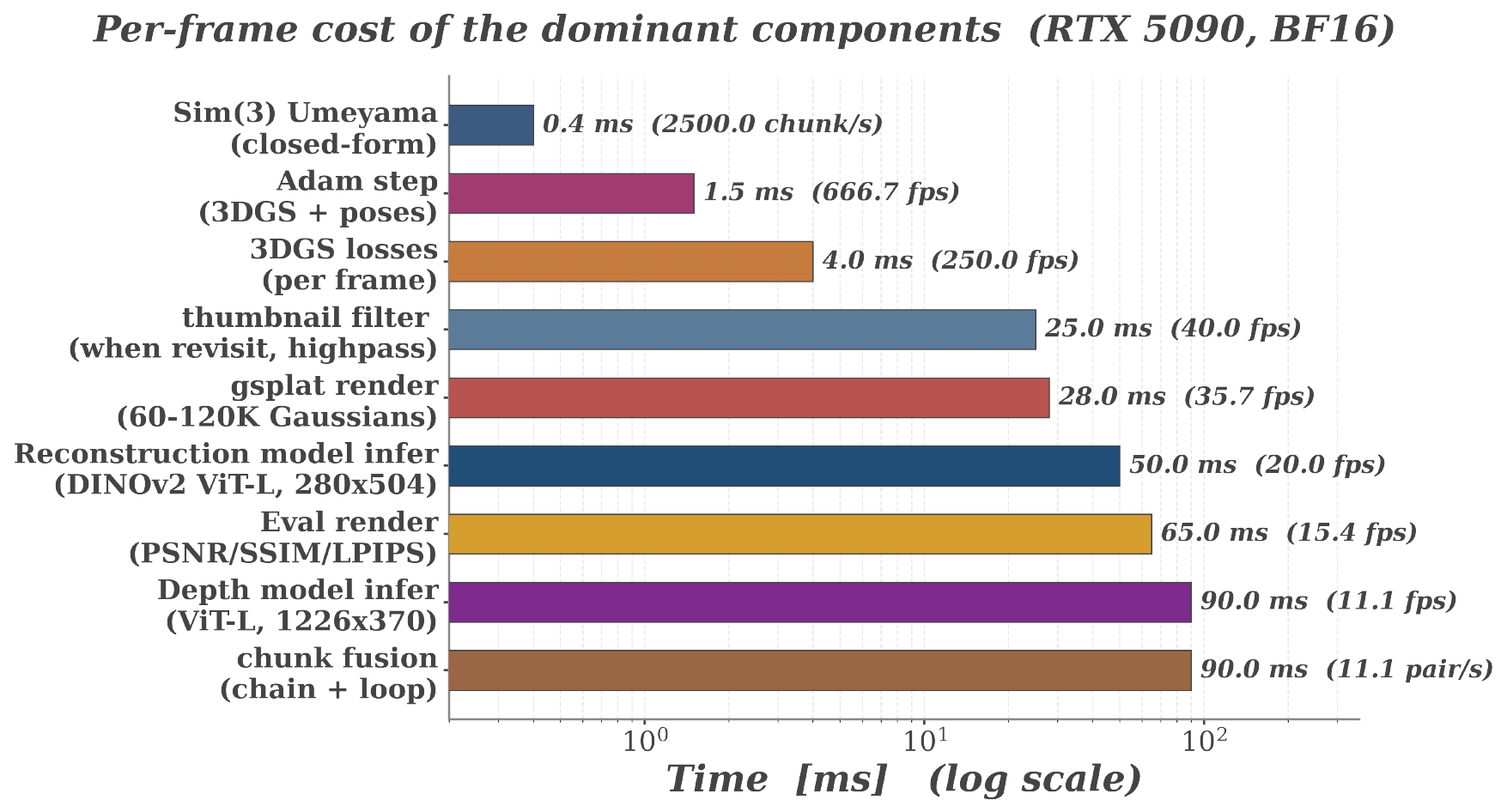}
    \caption{Per-frame computational cost of $\text{SG}^2$SLAM.}
    \label{fig:cost}
\end{figure}

\subsubsection{Joint Fusion and Optimization}

Every measurement collected above enters a single solve. Let $\mathcal{E}_{\mathrm{ch}}$ and $\mathcal{E}_{\mathrm{lp}}$ denote the chain and loop edge sets, and let $\hat{\mathbf{S}}_{ij}=\langle\hat{\sigma}_{ij},\hat{\mathbf{R}}_{ij},\hat{\mathbf{b}}_{ij}\rangle$ denote the similarity measured on edge $(i,j)$. With $\mathbf{G}_1$ fixed, the unknowns form a global parameter vector $\mathbf{p}$ over the remaining chunks,
\begin{equation}
\begin{aligned}
\mathbf{p}_k &= \big[\log\sigma_k,\ \boldsymbol{\omega}_k,\ \mathbf{b}_k\big]^{\top}
   \in\mathbb{R}^{7}, \quad k=2,\ldots,K,\\
\mathbf{p} &= \big[\mathbf{p}_2^{\top},\ldots,\mathbf{p}_K^{\top}\big]^{\top}
   \in\mathbb{R}^{7(K-1)},
\end{aligned}
\label{eq:params}
\end{equation}
where $\boldsymbol{\omega}_k$ is the rotation vector of $\mathbf{R}_k$, and the logarithm leaves the scale free of sign ambiguity. Define the predicted relative similarity $\mathbf{S}^{\mathrm{pred}}_{ij}$ and the map $\boldsymbol{\xi}$ that transforms similarity into seven-coordinate vectors for the optimizer to work with unconstrained parameters.
\vspace{2mm}
\begin{equation}
\small
\begin{aligned}
\mathbf{S}^{\mathrm{pred}}_{ij} &= \mathbf{G}_i^{-1}\mathbf{G}_j,\\
\boldsymbol{\xi}(\mathbf{S}) &= \big[\log\sigma_S,\ \big(\log\mathbf{R}_S\big)^{\vee},\
   \mathbf{t}_S\big]^{\top},
\end{aligned}
\label{eq:predrel}
\end{equation}
where $(\cdot)^{\vee}$ is the axis-angle vector of $\mathbf{R}_S$, and $(\sigma_S,\mathbf{R}_S,\mathbf{t}_S)$ are the similarity parameters of $\mathbf{S}$. The residual on edge $(i,j)\in\mathcal{E}$ whitens the discrepancy between prediction and measurement by the uncertainty of the measurement.
\begin{equation}
\label{eq:resid}
\mathbf{r}_{ij}=w_{ij}\,\boldsymbol{\Omega}_{ij}^{-1/2}\,
\boldsymbol{\xi}\big(\hat{\mathbf{S}}_{ij}^{-1}\,\mathbf{S}^{\mathrm{pred}}_{ij}\big),
\end{equation}
with $\boldsymbol{\Omega}_{ij}$ the diagonal covariance of the seven measurement coordinates and $w_{ij}>0$ is the uncertainty weight. Carrying $\log\sigma$ inside the residual allows a single loop edge to settle the scale error accumulated across every chain boundary preceding the revisit.

\begin{algorithm}[t]
\caption{$\text{SG}^2$SLAM pipeline.}
\label{alg:pipeline}
\footnotesize
\begin{algorithmic}[1]
\Require image stream $\{\mathcal{V}_t\}$, intrinsics $\mathbf{K}$
\Ensure trajectory $\{\mathbf{T}_t\}$ and map $\mathcal{M}$
\State $\mathcal{M}\leftarrow\emptyset$, $\mathcal{E}\leftarrow\emptyset$
\For{each chunk $k$ over frames $\mathcal{F}_k$, overlapping by $N_o$}
  \Statex \Comment{Sec.~\ref{ss:prior}: spatial geometry prior}
  \State $\Phi_\Theta(\mathcal{V}_t)\Rightarrow$
         $\{\hat{\boldsymbol{\Delta}}_i,\hat{\mathbf{X}}_{k,i},\hat{c}_{k,i}\}$
         \Comment{rays, poses, points}
  \State $\bar{s}_k\leftarrow\mathrm{med}_{\mathbf{u}}\,
         d^{\Psi}(\mathbf{u})/\hat{z}_i(\mathbf{u})$ \Comment{metric depth}
  \State $\mathbf{X}^{(m)}_{k,i}\leftarrow\bar{s}_k\,\hat{\mathbf{X}}_{k,i}$;
         $\{\mathbf{T}^{\mathcal{A},(m)}_{k,i}\}$
         \Comment{metric chunk}
  \Statex \Comment{Sec.~\ref{ss:backend}: ray-conditioned 3DGS}
  \State $\mathcal{K}_k\leftarrow$ confident frames
  \State $\mathcal{M}\leftarrow\mathcal{M}\cup\{\mathbf{X}^{(m)}_{k,i}\}_{i\in\mathcal{K}_k}$
         \Comment{anchored in $\mathcal{A}$}
  \State $(\mathcal{M},\{\mathbf{T}^{\mathcal{A},(m)}_{k,i}\})\leftarrow$
         bundle adjustment over $\mathcal{K}_k$
  \Statex \Comment{Sec.~\ref{ss:graph}: joint global fusion}
  \State $\hat{\mathbf{S}}_{k-1,k}\leftarrow$ overlap frames, anchor ratio
         
  \State $\hat{\mathbf{S}}_{ij}\leftarrow$ revisits, if any
  \State $\{\mathbf{G}_k\}_{k\geq2}\leftarrow$ pose graph over $\mathcal{E}$
  \State $\mathcal{M}\leftarrow$ re-express under $\{\mathbf{G}_k\}$;\;
         $\mathcal{E}\leftarrow\emptyset$
\EndFor
\State \Return $\{\mathbf{T}_t\}$, $\mathcal{M}$
\end{algorithmic}
\end{algorithm}

A loop edge measures scale directly from ranges and receives a high weight, while a chain edge inherits the scale error of the two chunks and receives a low weight, so the objective of the solve is linear.
\begin{equation}
\label{eq:graphcost}
\mathbf{p}^{\star}=\arg\min_{\mathbf{p}}\ \sum_{(i,j)\in\mathcal{E}} w_{ij}\,\mathbf{r}_{ij}^{\top}\mathbf{r}_{ij},
\end{equation}
where the minimization runs to convergence by Gauss-Newton iteration on global parameter vectors $\mathbf{p}$.

Finally, the solved placements return to the trajectory. Every frame $t$ is observed by a set of chunks $\mathcal{K}(t)$, and for each chunk $k$ covering $t$, the placed pose follows from $\mathbf{G}_k$. Let $r^{\mathrm{in}}_k$ be the number of frames that chunk $k$ shares with the previous chunk, and $r^{\mathrm{out}}_k$ the number shared with the next; let $n_k$ be the length of chunk $k$, and let $i(t,k)$ index frame $t$ within chunk $k$. The weight ramps across each overlap.
\begin{equation}
\small
\begin{aligned}
\gamma_{k}(t) &= \frac{i(t,k)+1/2}{r^{\mathrm{in}}_k}\cdot
   \frac{n_k-i(t,k)-1/2}{r^{\mathrm{out}}_k}, \\
w_{k}(t) &= \frac{\gamma_k(t)}
   {\sum_{k'\in\mathcal{K}(t)}\gamma_{k'}(t)},
\end{aligned}
\label{eq:fuseweight}
\end{equation}
with either factor set to unity when the corresponding overlap is empty. A frame covered by a single chunk takes the placed pose unchanged; a frame covered by several chunks takes the weighted mean of the scales and translations, and the weighted chordal mean of the rotations.
\begin{equation}
\small
\begin{aligned}
\sigma_t = \sum_{k} w_k\,\sigma_k, \quad \mathbf{b}_t = \sum_{k} w_k\,\mathbf{b}_k, \quad \mathbf{R}_t = \Big(\textstyle\sum_{k} w_k\,\mathbf{R}_k\Big)^{\dagger},
\end{aligned}
\label{eq:fuse}
\end{equation}
where $\sigma_k$, $\mathbf{R}_k$, and $\mathbf{b}_k$ are read from $\mathbf{G}_k$ in global coordinates. Spreading the transition across the full overlap keeps consecutive frames inside the motion envelope implied by the neighboring poses, resulting in a continuous fused trajectory. The overall pipeline is summarized in Alg. \ref{alg:pipeline}.

%% file: Exp.tex
\subsection{Datasets and Experimental Setup.}

Our experiments are conducted on the Nvidia RTX 5090 GPU (32 GB VRAM) with 200 streaming frames per chunk and 30 overlapping frames in the frontend. To evaluate our system in outdoor scenes, we use the KITTI and Waymo Open Dataset (WOD), which serve as standard benchmarks for large-scale monocular visual SLAM evaluation. For the KITTI Odometry Benchmark, we conduct experiments using the 11 training sequences (00–10), utilizing the monocular color camera stream. These sequences cover diverse suburban, highway, and urban environments, with highly accurate ground-truth trajectories \cite{geiger2012kitti}. From the WOD perception split, we select 9 representative sequences, utilizing the front-facing high-resolution camera to test our system against varying illumination, high dynamic traffic, and complex highway geometries \cite{sun2020waymo}. Additionally, we conduct tracking experiments without loop closure on our own driving sequence under an OMNIVISION OX03F10 automotive monocular camera, where the ground truth pose is recovered using a tightly-coupled sensor suite comprising a u-blox ZED-F9P multi-band RTK-GNSS receiver, a high-frequency IMU, a Velodyne VLP-32 3D LiDAR, and wheel encoders accessed via the vehicle CAN bus.

\begin{table}[t]
\caption{Rendering performance on KITTI \cite{geiger2012kitti} dataset. \\
LC = Loop Closure. \\ OOM = GPU out of memory, \quad \textbf{---} = tracking failure.}
\vspace{-3mm}
    \centering
 \resizebox{\linewidth}{!}{
    \begin{tabular}{c|c|ccccccccccc}
    \toprule
        \textbf{Methods} & \textbf{Metrics} & \textbf{00} & \textbf{01} & \textbf{02} & \textbf{03} & \textbf{04} & \textbf{05} & \textbf{06} & \textbf{07} & \textbf{08} & \textbf{09} & \textbf{10}   \\ \midrule
         & PSNR $\uparrow$ & 10.09 & 16.40 & 8.78 & 11.83 & 17.43 & 10.83 & 12.66 & 8.69 & 6.90 & 9.71 & 8.66  \\ 
        MonoGS \cite{matsuki2024gaussian} & SSIM $\uparrow$ & 0.43 & 0.58 & 0.31 & 0.36 & 0.55 & 0.37 & 0.43 & 0.34 & 0.27 & 0.27 & 0.32  \\ 
          & LPIPS $\downarrow$ & 0.82 & 0.66 & 0.85 & 0.82 & 0.55 & 0.84 & 0.76 & 0.86 & 0.89 & 0.81 & 0.84  \\ 
        \midrule
        
         & PSNR $\uparrow$ & 6.96 & 13.94 & \textbf{---} & 14.95 & 17.90 & \textbf{---} & 12.68 & \textbf{---} & \textbf{---} & 11.03 & 12.82 \\ 
        S3PO-GS \cite{sandstrom2024splat} & SSIM $\uparrow$ & 0.64 & 0.55 & \textbf{---} & 0.39 & 0.54 & \textbf{---} & 0.44 & \textbf{---} & \textbf{---} & 0.33 & 0.41  \\ 
          & LPIPS $\downarrow$ & 0.90 & 0.68 & \textbf{---} & 0.73 & 0.57 & \textbf{---} & 0.75 & \textbf{---} & \textbf{---} & 0.83 & 0.78   \\ 
        \midrule
        
        & PSNR $\uparrow$ & OOM & 15.68 & OOM & 16.26 & 15.72 & 14.98 & 14.96 & 14.83 & OOM & 15.83 & 16.09 \\ 
        DiskChunGS \cite{sandstrom2024splat} & SSIM $\uparrow$ & OOM & 0.80 & OOM & 0.77 & 0.68 & 0.76 & 0.79 & 0.74 & OOM & 0.78 & 0.79  \\ 
          & LPIPS $\downarrow$ & OOM & 0.82 & OOM & 0.79 & 0.73 & 0.81 & 0.83 & 0.79 & OOM & 0.78 & 0.78   \\ 
        \midrule
        
        & PSNR $\uparrow$ & OOM & 23.69 & OOM & 23.03 & 23.87 & 21.89 & 23.09 & 22.46 & OOM & 21.95 & 21.48 \\ 
        GigaSLAM \cite{sandstrom2024splat} & SSIM $\uparrow$ & OOM & 0.80 & OOM & 0.81 & \textbf{0.90} & 0.85 & 0.86 & 0.83 & OOM & 0.88 & 0.82  \\ 
          & LPIPS $\downarrow$ & OOM & 0.50 & OOM & 0.43 & 0.40 & \textbf{0.46} & 0.49 & 0.40 & OOM & 0.45 & 0.42   \\ 
        \midrule
        
         & PSNR $\uparrow$ & \textbf{22.66} & \textbf{23.83} & \textbf{20.32} & \textbf{23.48} & \textbf{23.99} & \textbf{22.56} & \textbf{23.18} & \textbf{22.83} & \textbf{22.57} & \textbf{22.03} & \textbf{22.27}  \\ 
       \textbf{Ours} & SSIM $\uparrow$ & \textbf{0.82} & \textbf{0.86} & \textbf{0.79} & \textbf{0.89} & 0.89 & \textbf{0.86} & \textbf{0.88} & \textbf{0.89} & \textbf{0.83} & \textbf{0.88} & \textbf{0.85}  \\ 
        ~ & LPIPS $\downarrow$ & \textbf{0.36} & \textbf{0.43} & \textbf{0.39} & \textbf{0.40} & \textbf{0.40} & 0.48 & \textbf{0.46} & \textbf{0.38} & \textbf{0.39} & \textbf{0.44} & \textbf{0.40} \\ \bottomrule
    \end{tabular}
      }
  \label{table:kitti_render}
  \vspace{-5mm}
\end{table}

We report our backend mapping performance through the rendering quality of 3DGS checkpoints on the keyframes in Tab. \ref{table:kitti_render}. Extensive outdoor driving 3DGS mapping experiments are conducted on the challenging KITTI dataset.

\begin{table}[h]
\caption{Tracking Performance on KITTI~\cite{geiger2012kitti} dataset (ATE RMSE in meters $\downarrow$). \\
LC = Loop Closure. \\ OOM = GPU out of memory, \quad \textbf{---} = tracking failure.}
\vspace{-3mm}
\setlength{\tabcolsep}{3pt} 
\renewcommand{\arraystretch}{0.9}
\begin{centering}
\begin{tabular}{lcccccc}
\hline 
\textbf{Seq.} & LC & \textbf{MonoGS} & \textbf{S3PO-GS} & \textbf{DiskChunGS} & \textbf{GigaSLAM} & \textbf{Ours} \tabularnewline
\hline 
\textbf{00} & \textcolor{green}{$\checkmark$} & 200.92 & 168.97 & OOM & OOM & \textbf{5.42} \tabularnewline
\textbf{01} & \textcolor{red}{$\times$} & 436.78 & 214.88 & 208.82 & 209.17 & \textbf{22.34} \tabularnewline
\textbf{02} & \textcolor{green}{$\checkmark$} & \textbf{---} & \textbf{---} & OOM & OOM & \textbf{9.83} \tabularnewline
\textbf{03} & \textcolor{red}{$\times$} & 25.63 & 18.75 & 1.34 & \textbf{1.33} & 1.77 \tabularnewline
\textbf{04} & \textcolor{red}{$\times$} & 18.65 & 5.99 & 1.72 & 1.69 & \textbf{1.65} \tabularnewline
\textbf{05} & \textcolor{green}{$\checkmark$} & \textbf{---} & \textbf{---} & 4.85 & \textbf{4.54} & 5.09 \tabularnewline
\textbf{06} & \textcolor{green}{$\checkmark$} & 109.30 & 69.18 & 1.37 & 1.41 & \textbf{1.30} \tabularnewline
\textbf{07} & \textcolor{green}{$\checkmark$} & \textbf{---} & \textbf{---} & 2.82 & 2.67 & \textbf{1.33} \tabularnewline
\textbf{08} & \textcolor{green}{$\checkmark$} & \textbf{---} & \textbf{---} & OOM & OOM & \textbf{12.21} \tabularnewline
\textbf{09} & \textcolor{green}{$\checkmark$} & \textbf{---} & 118.09 & 4.74 & \textbf{4.70} & 5.68 \tabularnewline
\textbf{10} & \textcolor{red}{$\times$} & \textbf{---} & 41.05 & 2.11 & 2.14 & \textbf{1.98} \tabularnewline

\hline 
\end{tabular}
\par\end{centering}
\label{tab:kitti_tracking_ate}
\vspace{-3mm}
\end{table}

\subsection{Experimental Results}

\begin{figure}[h]
    \centering
    \begin{subfigure}[b]{0.156\textwidth}
        \includegraphics[width=\textwidth]{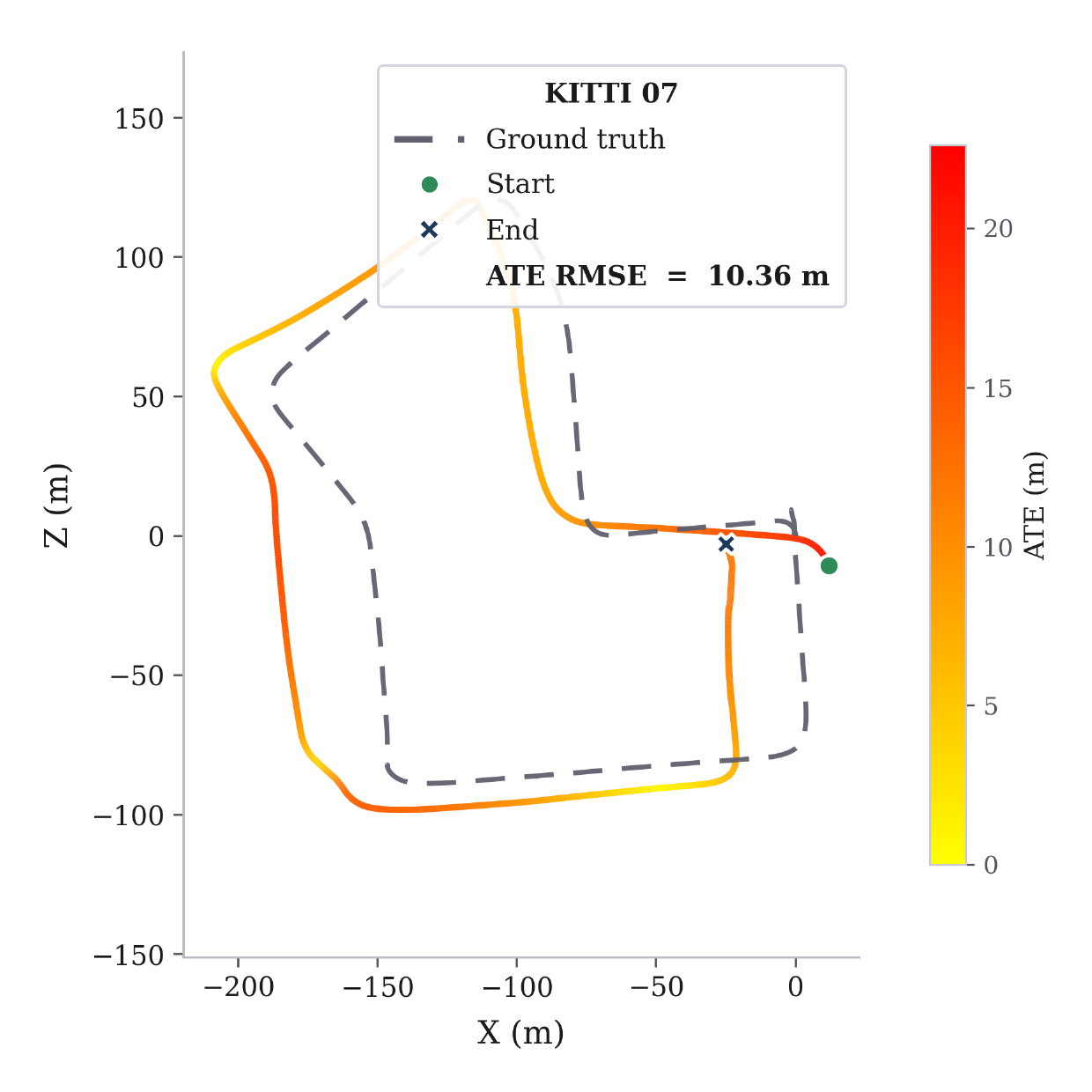}
    \end{subfigure}
    \hfill
    \begin{subfigure}[b]{0.156\textwidth}
        \includegraphics[width=\textwidth]{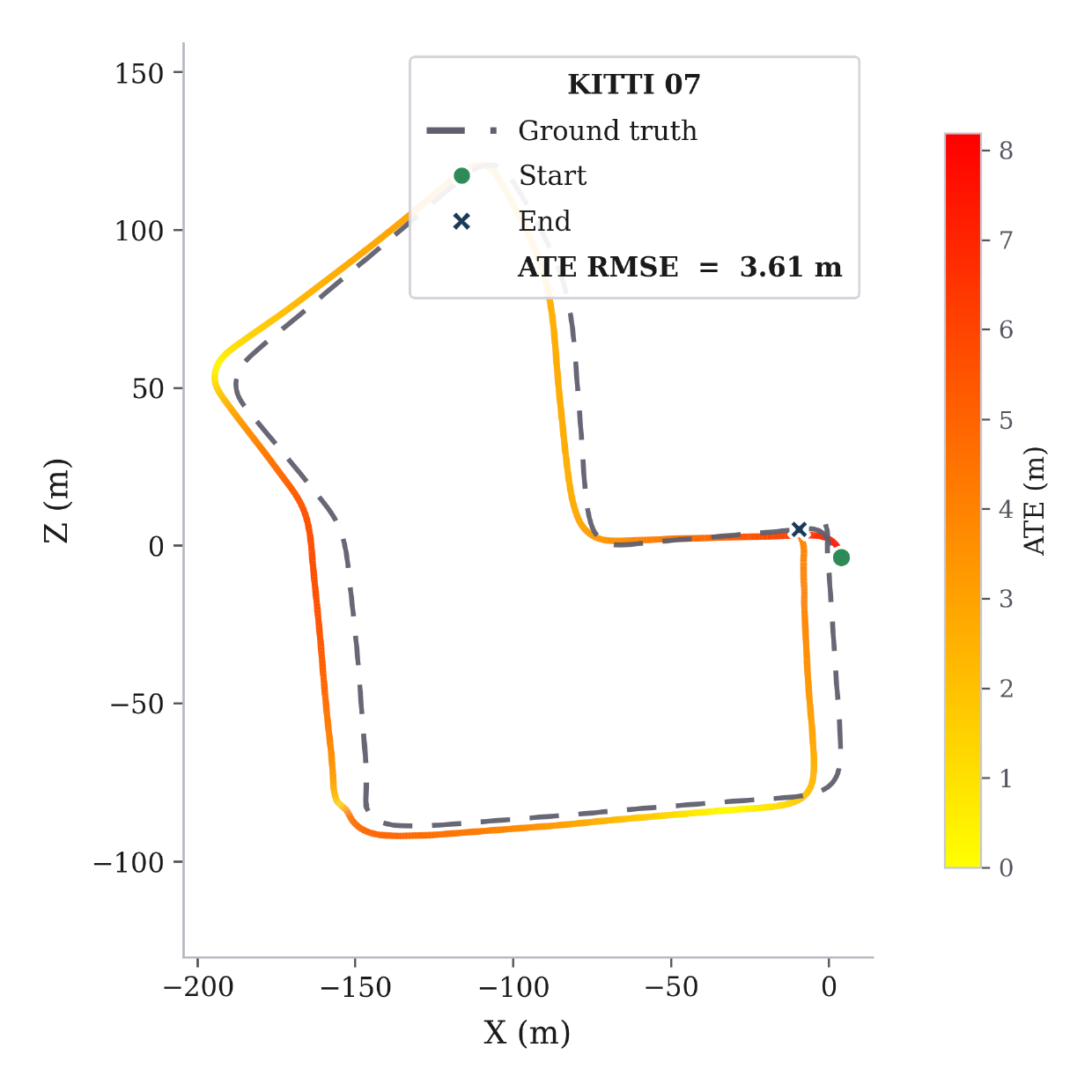}
    \end{subfigure}
    \hfill
    \begin{subfigure}[b]{0.156\textwidth}
        \includegraphics[width=\textwidth]{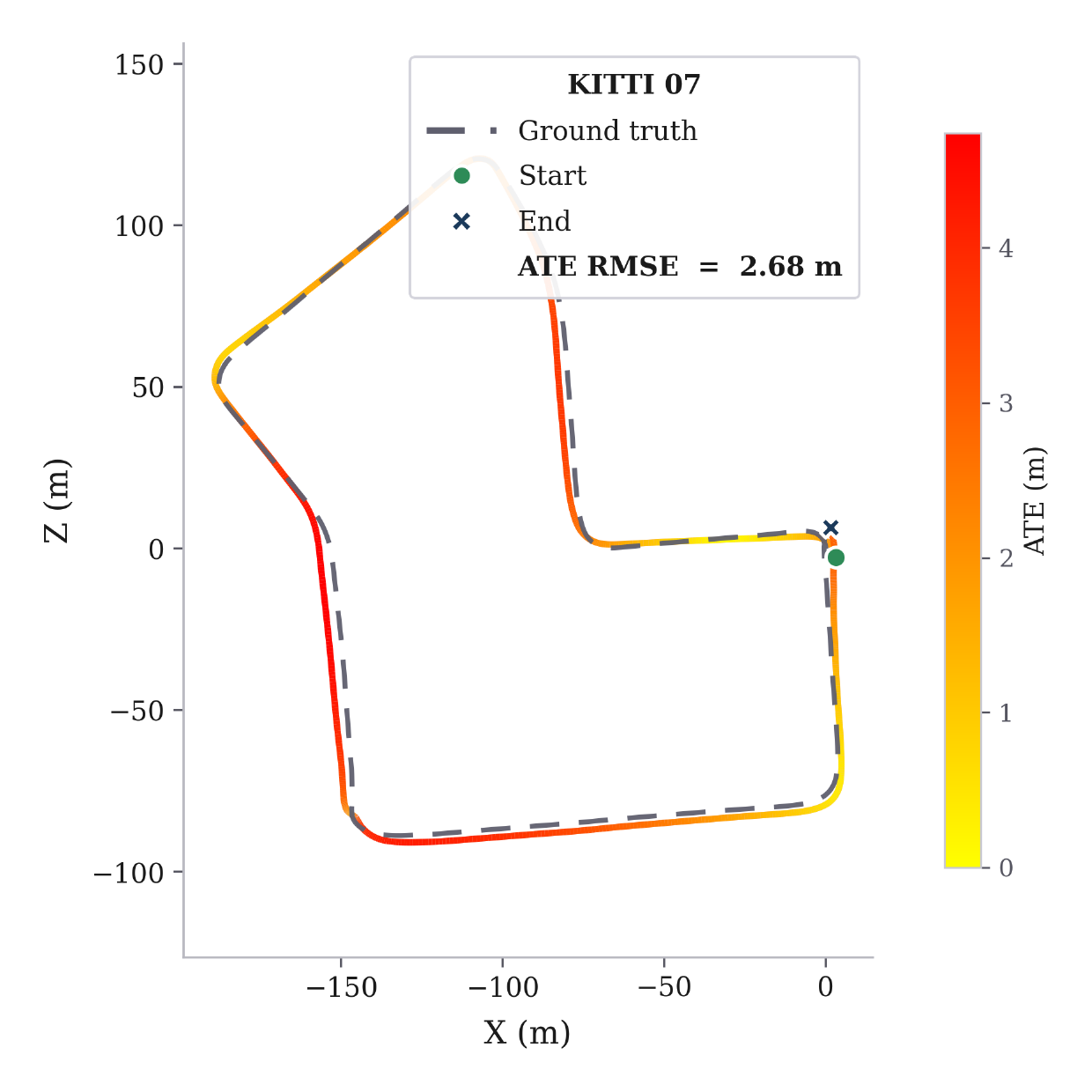}
    \end{subfigure}

    \vspace{1ex} 

    \begin{subfigure}[b]{0.156\textwidth}
        \includegraphics[width=\textwidth]{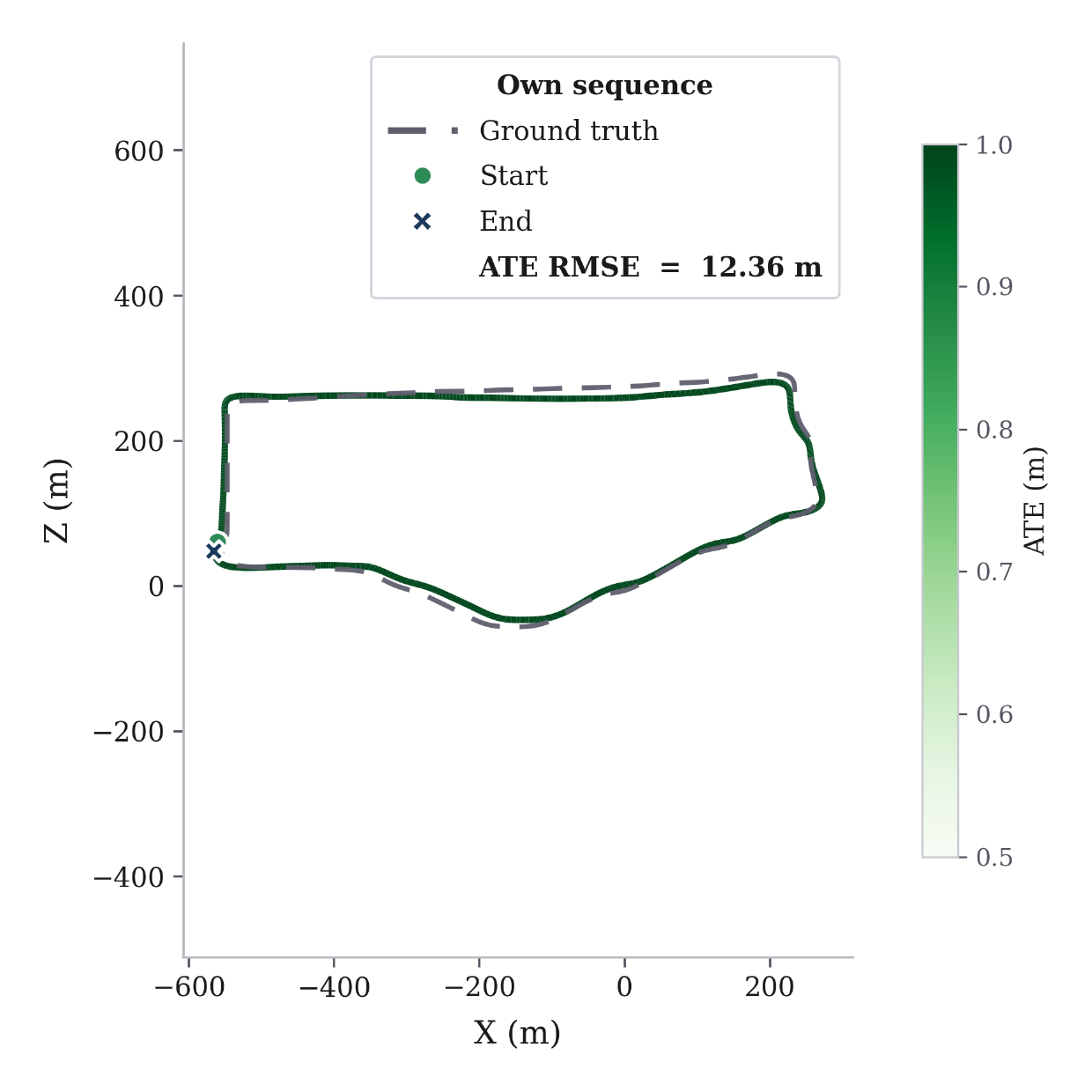}
        \caption{DiskChunkGS}
    \end{subfigure}
    \hfill
    \begin{subfigure}[b]{0.156\textwidth}
        \includegraphics[width=\textwidth]{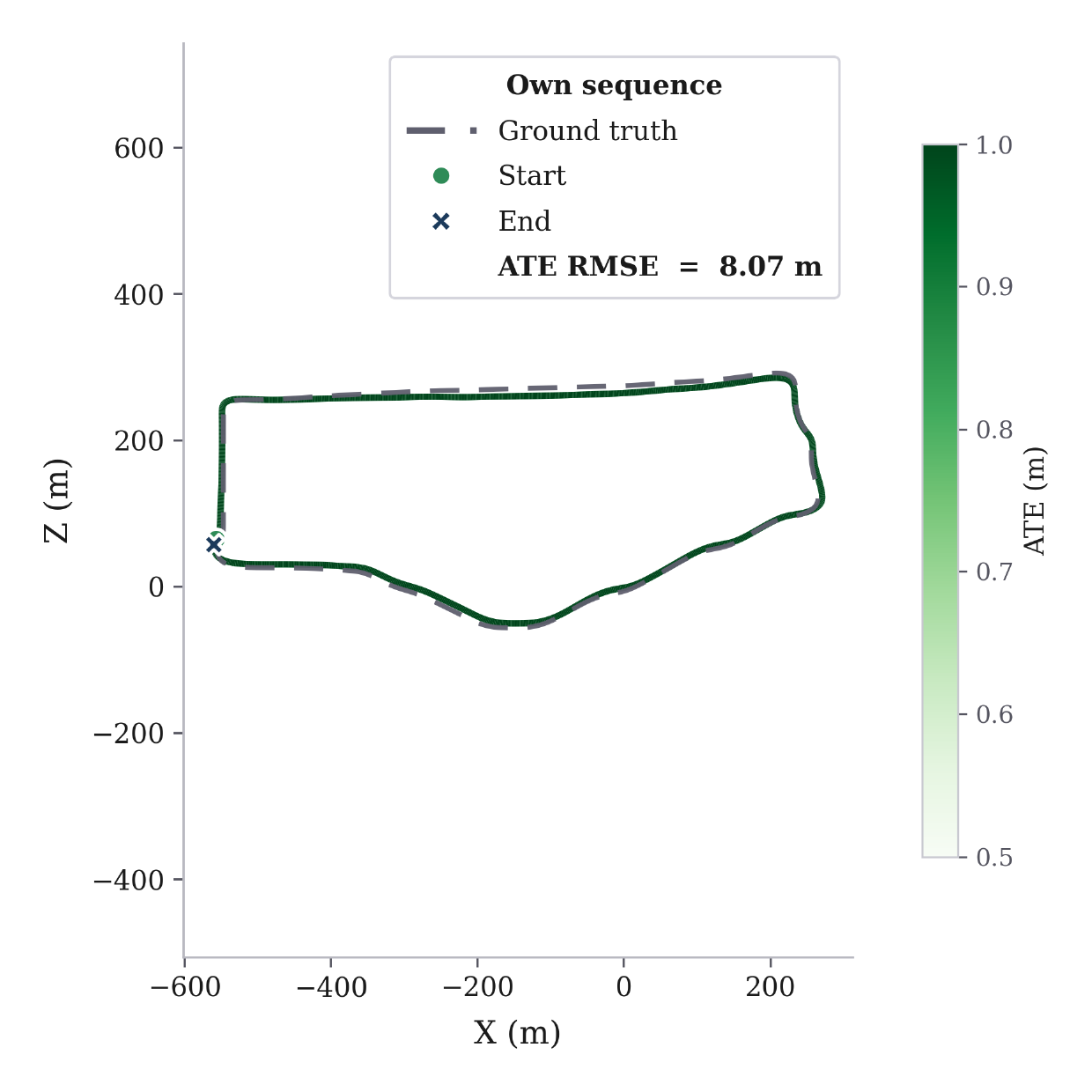}
        \caption{GigaSLAM}
    \end{subfigure}
    \hfill
    \begin{subfigure}[b]{0.156\textwidth}
        \includegraphics[width=\textwidth]{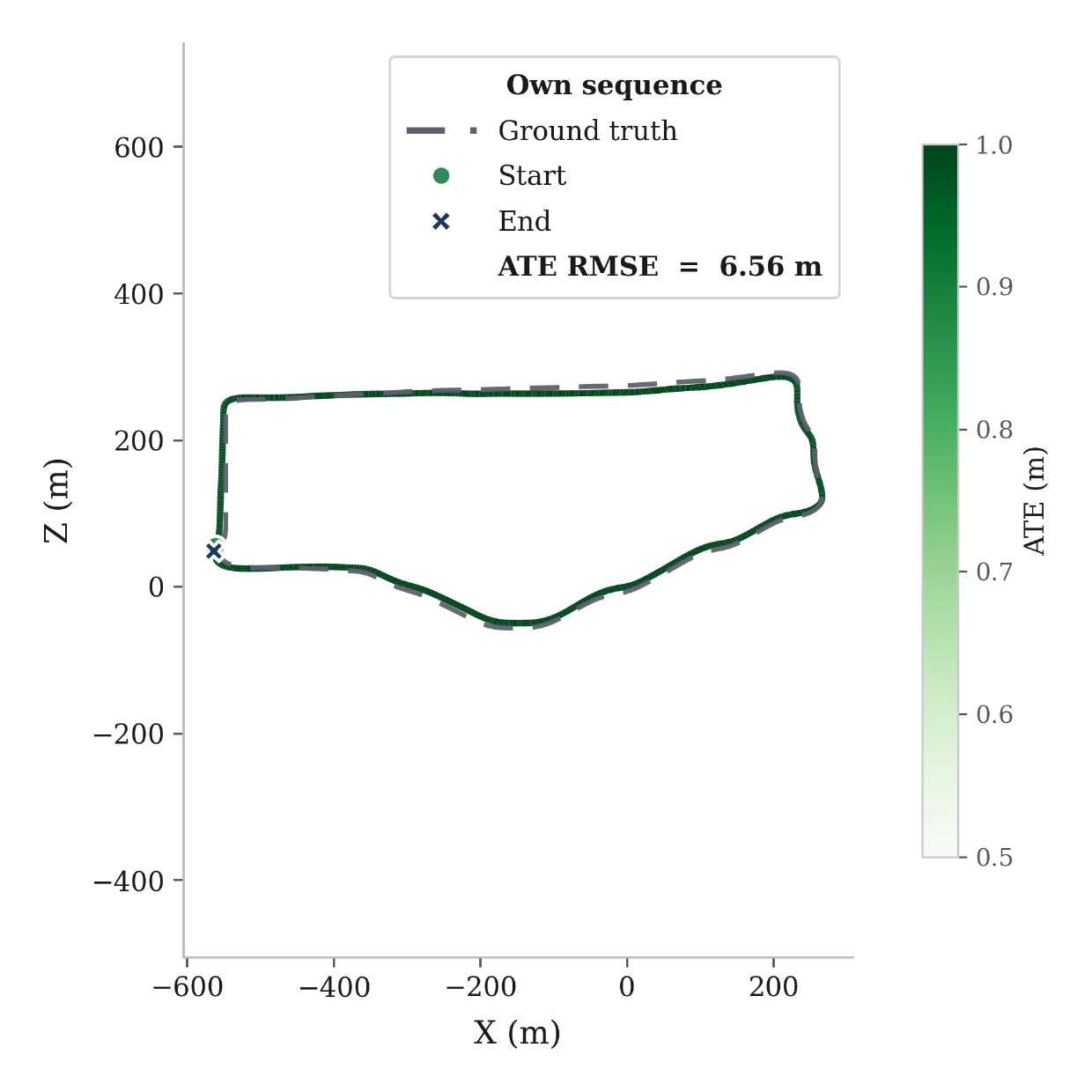}
        \caption{\textbf{Ours}}
    \end{subfigure}

    \caption{Tracking comparison on KITTI dataset and our own sequence without loop closure.}
    \label{fig:2x3_grid}
\end{figure}

Following standard monocular SLAM evaluation protocols, the estimated trajectories are aligned with the ground truth using a Sim(3) Umeyama alignment to resolve the inherent monocular scale ambiguity. System performance is quantitatively measured using the Absolute Trajectory Error (ATE RMSE) in meters to evaluate overall global consistency after scale correction. In Tab. \ref{tab:kitti_tracking_ate} and Tab. \ref{tab:waymo_tracking_ate}, our $\text{SG}^2$SLAM surpasses the existing baselines in the outdoor navigation task based on the chunking inference of the depth and geometry model, with an average ATE RMSE of $6.24$ m on 11 KITTI sequences and $0.63$ m on 9 WOD sequences. In Fig. \ref{fig:2x3_grid}, we show the comparison of tracking performance without loop closure, where our $\text{SG}^2$SLAM achieves an average ATE of $4.62$ m on KITTI Sequence 07 and our own real-world driving sequence, exceeding GigaSLAM by $21\%$ because the streaming prior provides the backend with the spatial geometry in chunks, enabling the backend's global solve to be well-conditioned, while a per-frame 2D–3D association incorporates the tracking error of every frame.

\begin{table}[t]
\caption{Tracking Performance on Waymo Open Dataset~\cite{sun2020waymo} dataset (ATE RMSE in meters $\downarrow$). \\
OOM = GPU out of memory, \quad \textbf{---} = tracking failure.}
\vspace{-3mm}
\setlength{\tabcolsep}{3pt} 
\renewcommand{\arraystretch}{1.0}
\begin{centering}
\begin{tabular}{lccccc}
\hline 
\textbf{Seq.} & \textbf{MonoGS} & \textbf{S3PO-GS} & \textbf{DiskChunGS} & \textbf{GigaSLAM} & \textbf{Ours} \tabularnewline
\hline 
\textbf{100613} & 6.96 & 0.66 & 0.30 & 0.36 & \textbf{0.18} \tabularnewline
\textbf{106762} & 18.01 & 0.61 & \textbf{0.12} & 0.43 & 0.20 \tabularnewline
\textbf{132384} & 12.04 & 1.54 & \textbf{0.05} & 1.23 & 0.67 \tabularnewline
\textbf{13476} & 3.30 & \textbf{---} & \textbf{0.45} & 3.18 & 0.92 \tabularnewline
\textbf{152706} & 9.30 & 0.96 & 0.29 & 0.63 & \textbf{0.22} \tabularnewline
\textbf{153495} & 5.99 & 2.03 & \textbf{---} & 0.73 & \textbf{0.55} \tabularnewline
\textbf{158686} & 8.40 & 1.04 & 0.64 & 0.66 & \textbf{0.61} \tabularnewline
\textbf{163454} & 11.21 & 0.95 & \textbf{0.55} & 5.11 & 1.60 \tabularnewline
\textbf{405841} & 1.63 & \textbf{0.49} & 0.50 & 0.79 & 0.68 \tabularnewline

\hline 
\end{tabular}
\par\end{centering}
\label{tab:waymo_tracking_ate}
\vspace{-3mm}
\end{table}

On 3DGS rendering quality, we report the Peak Signal-to-Noise Ratio (PSNR), Structural Similarity Index Measure (SSIM), and Learned Perceptual Image Patch Similarity (LPIPS) from the baselines and our system, without post color refinement, on the challenging KITTI dataset with a panoramic aspect ratio in Tab. \ref{table:kitti_render}, where our $\text{SG}^2$SLAM reaches an average PSNR of $22.70$ dB, an average SSIM of $0.86$, and an average LPIPS of $0.41$.

Additionally, in Fig. \ref{fig:cost}, we show our system's per-frame computational cost comparison on a logarithmic scale, where the chunk fusion loop in the backend dominates wall-clock time; the reconstruction model and metric depth model set the per-frame frontend budget.

\subsection{Qualitative Results}

We show the visual comparison of the 3DGS rendered RGB image and depth image in Fig. \ref{fig:3x5_grid} and Fig. \ref{fig:3x3_grid} from the challenging KITTI dataset without post color refinement after running the tracking and mapping. Our rendering quality displays the highest fidelity in photorealistic and depth information.

\begin{table}[h]
\vspace{-2mm}
\centering
\caption{Ablation study on our design for the main modules and the effects of different setups.}
\vspace{-3mm}
\label{table:abla}
\resizebox{0.48\textwidth}{!}{
\begin{tabular}{c|c c c c}
\toprule
     Ablations & ATE RMSE $\downarrow$ & PSNR $\uparrow$ & SSIM $\uparrow$ & LPIPS $\downarrow$ \\ \hline
    
    Vanilla 3DGS & 1.49 & 22.12 & 0.78 & 0.50 \\
    Our 3DGS & \textbf{1.49} & \textbf{23.41} & \textbf{0.89} & \textbf{0.39} \\
    \midrule
    Pair-wise Frontend & 2.65 & 19.95 & 0.67 & 0.59 \\
    Streaming Frontend & \textbf{1.49} & \textbf{23.41} & \textbf{0.89} & \textbf{0.39} \\
    \midrule
    200 frames / chunk & \textbf{1.49} & \textbf{23.41} & \textbf{0.89} & \textbf{0.39} \\
    300 frames / chunk & 1.92 & 22.42 & 0.80 & 0.43 \\
    500 frames / chunk & 2.42 & 21.51 & 0.79 & 0.49 \\
    
    \bottomrule
\end{tabular}}
\vspace{-5mm}
\end{table}

\subsection{Ablation Study}

In Tab. \ref{table:abla}, we show the ablation results of our design for the main modules and the effects of different setups on tracking and rendering performance on the KITTI dataset, where we report the average metrics across Sequence 04 and Sequence 07. Our ray-conditioned 3DGS kernel improves mapping performance under monocular SLAM, as demonstrated by the improvements in PSNR, SSIM, and LPIPS. For the pair-wise frontend experiment, we feed the images in pairs to MASt3R \cite{leroy2024mast3r} to obtain the estimated relative poses, while different frames in chunks determine the noise level of the fusion anchor and the initialization of 3DGS training. The ablations show that more tracking errors may accumulate frame-wise along the run, and the design of the chunk size affects both the frontend and backend due to how much prior noise propagates throughout the fusion and mapping.